\documentclass[lettersize,journal]{IEEEtran}

\usepackage{amsmath,amsfonts, amssymb}
\usepackage{algorithm}
\usepackage[noend]{algpseudocode}
\usepackage{array}
\usepackage{balance}
\usepackage{boldline}
\usepackage[font=scriptsize,labelfont=bf]{caption}
\usepackage[nomath]{cellspace}
\makeatletter
\let\@startpbox@action\@startpbox   
\makeatother
\usepackage{duckuments}
\usepackage{float}
\usepackage{graphicx}
\usepackage{multicol}
\usepackage{multirow}
\usepackage[numbers,sort&compress]{natbib}
\usepackage{overpic}
\usepackage{stfloats}
\usepackage[font=scriptsize,labelfont=bf]{subcaption}
\usepackage{tabularx}
\usepackage{titlesec}
\usepackage{url}
\usepackage{verbatim}
\usepackage[dvipsnames, table]{xcolor}

\usepackage[bookmarks=true]{hyperref}
\usepackage{cleveref}
\hypersetup{
    colorlinks=true,
    linkcolor=blue,
    filecolor=magenta,      
    urlcolor=cyan,
    pdfpagemode=FullScreen,
}

\titlespacing\subsection{0pt}{6pt plus 2pt minus 2pt}{2pt plus 2pt minus 2pt}
\titleformat{\subsubsection}[runin]
   {\itshape}
   {\thesubsubsectiondis}
   {0.5em}
   {}
   [:]
\titlespacing\subsubsection{\parindent}{0pt}{0.5em}

\usepackage{stix}
\usepackage{tikz}
\usepackage{marvosym}
\usepackage{fontawesome} 
\DeclareRobustCommand{\thickX}{
    \begin{tikzpicture}[baseline=0ex, line width=2, scale=0.13];
    \draw (0,0) -- (1,1);
    \draw (0,1) -- (1,0);
    \end{tikzpicture}}

\newcommand{\SymRIMESA}{\textcolor{sns:blue}{$\bigstar$}}
\newcommand{\SymCentralized}{\textcolor{sns:gray}{$\mdblkcircle$}}
\newcommand{\SymIMESA}{\textcolor{sns:light_blue}{$\pentagonblack$}}
\newcommand{\SymGNC}{\textcolor{sns:brown}{$\hexagonblack$}}
\newcommand{\SymPCM}{\textcolor{sns:dark:orange}{\Octosteel}}

\newcommand{\SymKIMESA}{\textcolor{sns:orange}{$\mdblksquare$}}
\newcommand{\SymDLGBP}{\textcolor{sns:red}{$\blacktriangle$}}
\newcommand{\SymDDFSAM}{\textcolor{sns:green}{$\thickX$}}
\newcommand{\SymIndep}{\textcolor{sns:pink}{$\blacklozenge$}}

\usepackage{amsfonts}
\usepackage{amsthm}

\newcommand{\mat}[1]{\begin{bmatrix}#1\end{bmatrix}}

\newcommand\blfootnote[1]{
    \begingroup
    \renewcommand\thefootnote{}\footnote{#1}
    \addtocounter{footnote}{-1}
    \endgroup
}
\newtheoremstyle{dense}
  {3pt} 
  {3pt} 
  {\itshape} 
  {} 
  {\bfseries} 
  {:} 
  {.5em} 
  {} 
\theoremstyle{dense}

\newtheorem{remark}{Remark}
\AtBeginEnvironment{remark}{\small}
\AtBeginEnvironment{quote}{\par\singlespacing\small}

\definecolor{tab:blue}{RGB}{31,119,180}
\definecolor{tab:orange}{RGB}{255,127,14}
\definecolor{tab:green}{RGB}{44,160,44}
\definecolor{tab:red}{RGB}{214,39,40}
\definecolor{tab:purple}{RGB}{148,103,189}

\definecolor{sns:blue}{rgb}{0.00392156862745098, 0.45098039215686275, 0.6980392156862745}
\definecolor{sns:orange}{rgb}{0.8705882352941177, 0.5607843137254902, 0.0196078431372549}
\definecolor{sns:green}{rgb}{0.00784313725490196, 0.6196078431372549, 0.45098039215686275}
\definecolor{sns:red}{rgb}{0.8352941176470589, 0.3686274509803922, 0.0}
\definecolor{sns:purple}{rgb}{0.8, 0.47058823529411764, 0.7372549019607844}
\definecolor{sns:brown}{rgb}{0.792156862745098, 0.5686274509803921, 0.3803921568627451}
\definecolor{sns:pink}{rgb}{0.984313725490196, 0.6862745098039216, 0.8941176470588236}
\definecolor{sns:gray}{rgb}{0.5803921568627451, 0.5803921568627451, 0.5803921568627451}
\definecolor{sns:yellow}{rgb}{0.9254901960784314, 0.8823529411764706, 0.2}
\definecolor{sns:light_blue}{rgb}{0.33725490196078434, 0.7058823529411765, 0.9137254901960784}

\definecolor{sns:dark:orange}{rgb}{0.6941176470588235, 0.25098039215686274, 0.050980392156862744}
\definecolor{sns:dark:blue}{rgb}{0.0, 0.10980392156862745, 0.4980392156862745}
\definecolor{sns:dark:light_blue}{rgb}{0.0, 0.38823529411764707, 0.4549019607843137}

\definecolor{sns:bright:blue}{rgb}{0.00784313725490196, 0.24313725490196078, 1.0}
\definecolor{sns:bright:light_blue}{rgb}{0.0, 0.8431372549019608, 1.0}

\definecolor{header:gray}{rgb}{0.6666666, 0.6666666, 0.6666666}
\definecolor{hls3:biased_prior}{rgb}{0.7176470588235294, 0.3411764705882353, 0.8588235294117647}
\definecolor{hls3:robot1}{rgb}{0.8588235294117647, 0.37254901960784315, 0.3411764705882353}
\definecolor{hls3:robot2}{rgb}{0.3411764705882353, 0.8588235294117647, 0.37254901960784315}
\definecolor{hls3:robot3}{rgb}{0.37254901960784315, 0.3411764705882353, 0.8588235294117647}

\DeclareMathOperator*{\argmax}{arg\,max}
\DeclareMathOperator*{\argmin}{arg\,min}

\newcommand{\logmap}[1]{\mathrm{Log}\left(#1\right)}

\newcommand{\norm}[1]{\left\|#1\right\|}

\newcommand{\inner}[2]{\left\langle #1, #2 \right\rangle}

\newcommand{\Ac}{\mathcal{A}}

\newcommand{\Cc}{\mathcal{C}}

\newcommand{\Ec}{\mathcal{E}}
\newcommand{\Fc}{\mathcal{F}}
\newcommand{\Gc}{\mathcal{G}}

\newcommand{\Ic}{\mathcal{I}}
\newcommand{\Jc}{\mathcal{J}}
\newcommand{\Kc}{\mathcal{K}}
\newcommand{\Lc}{\mathcal{L}}
\newcommand{\Mc}{\mathcal{M}}

\newcommand{\Oc}{\mathcal{O}}

\newcommand{\Qc}{\mathcal{Q}}
\newcommand{\Rc}{\mathcal{R}}

\newcommand{\Wc}{\mathcal{W}}

\newcommand{\Yc}{\mathcal{Y}}
\newcommand{\Zc}{\mathcal{Z}}

\newcommand{\Hfr}{\mathfrak{H}}

\newcommand{\Rfr}{\mathfrak{R}}
\newcommand{\Sfr}{\mathfrak{S}}

\newcommand{\Rb}{\mathbb{R}}

\begin{document}

\newcommand{\IEEEcopyrightnotice}{%
    This article has been accepted for publication in
    \emph{IEEE Transactions on Robotics}. DOI: 10.1109/TRO.2026.3706569
    \copyright~2026 IEEE.  Personal use of this material is permitted.  Permission from IEEE must be obtained for all other uses, in any current or future media, including reprinting/republishing this material for advertising or promotional purposes, creating new collective works, for resale or redistribution to servers or lists, or reuse of any copyrighted component of this work in other works.
}

\makeatletter
\def\IEEEtitletopspaceextra{0\baselineskip}
\def\ps@IEEEtitlepagestyle{%
  \def\@oddhead{%
    \vbox to \headheight{%
      \hbox to \textwidth{\hbox{}\@IEEEheaderstyle\leftmark\hfil\thepage}%
      \vskip 4pt
      \hbox to \textwidth{\parbox{\textwidth}{\scriptsize\IEEEcopyrightnotice}}%
      \vss}}%
  \let\@evenhead\@oddhead
  \let\@oddfoot\@empty
  \let\@evenfoot\@empty}
\makeatother


\title{riMESA: Consensus ADMM for Real-World Collaborative SLAM}
\author{Daniel McGann and Michael Kaess}

\markboth{IEEE TRANSACTIONS ON ROBOTICS, VOL. 42, 2026}{}

\twocolumn[{%
    \renewcommand\twocolumn[1][]{#1}
	\maketitle
    \vspace{-1.2cm}
    \begin{center}
        \centering
        \includegraphics[width=1\linewidth]{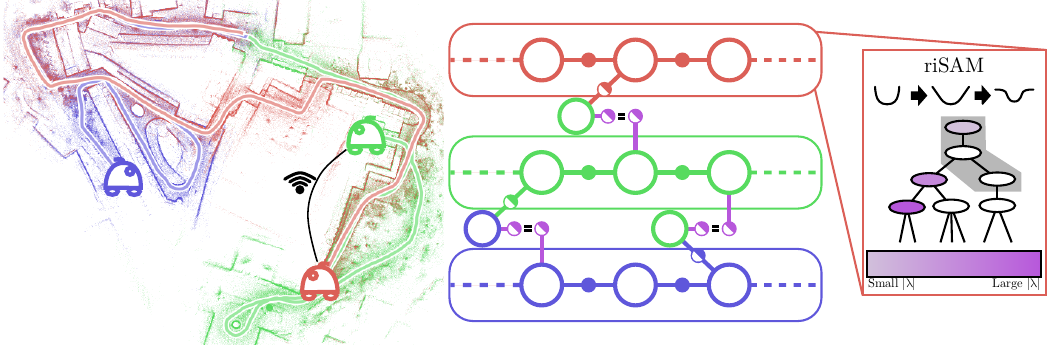}
        \vspace{-16pt}
        \captionof{figure}{An illustration of riMESA operating on real-world data (\texttt{kth\_r3\_00\_proradio}). riMESA estimates the state of a multi-robot team (\textcolor{hls3:robot1}{$\mdwhtcircle$},\textcolor{hls3:robot2}{$\mdwhtcircle$},\textcolor{hls3:robot3}{$\mdwhtcircle$}) from noisy, potentially incorrect measurements (\textcolor{hls3:robot1}{$\smblkcircle$},\textcolor{hls3:robot2}{$\smblkcircle$},\textcolor{hls3:robot3}{$\smblkcircle$}) using only sparse, unreliable communication (\faWifi). riMESA is a C-ADMM-based distributed optimization algorithm in which robots locally constrain shared state using ``biased priors'' (\textcolor{hls3:biased_prior}{$\smblkcircle$}). Over time, as communication is available, riMESA tightens equality constraints with dual variables ($\lambda$) to provide consistent solutions for the team. Meanwhile, robots incorporate new measurements efficiently using the riSAM algorithm, which handles potential outlier measurements (\rotatebox[origin=c]{45}{$\circlerighthalfblack$}) using M-Estimations and an incremental version of Graduated Non-Convexity, which efficiently updates only the relevant subproblem (\textcolor{header:gray}{$\mdblksquare$}) at each timestep.}
        \label{fig:rimesa-header-figure}
        \vspace{-6pt}
    \end{center}    
}]


\begin{abstract}
Collaborative Simultaneous Localization and Mapping (C-SLAM) is a fundamental capability for multi-robot teams as it enables downstream tasks like planning and navigation. However, existing C-SLAM back-end algorithms that are required to solve this problem struggle to address the practical realities of real-world deployments (i.e. communication limitations, outlier measurements, and online operation). In this paper we propose \textit{R}obust \textit{I}ncremental \textit{M}anifold \textit{E}dge-based \textit{S}eparable \textit{A}DMM (riMESA) -- a robust, incremental, and distributed C-SLAM back-end that is resilient to outliers, reliable in the face of limited communication, and can compute accurate state estimates for a multi-robot team in real-time. Through the development of riMESA, we, more broadly, make an argument for the use of Consensus Alternating Direction Method of Multipliers (C-ADMM) as a theoretical foundation for distributed optimization tasks in robotics like C-SLAM due to its flexibility, accuracy, and fast convergence. We conclude this work with an in-depth evaluation of riMESA on a variety of C-SLAM problem scenarios and communication network conditions using both synthetic and real-world C-SLAM data. These experiments demonstrate that riMESA is able to generalize across conditions, produce accurate state estimates, operate in real-time, and outperform the accuracy of prior works by a factor >7x on real-world datasets.
\end{abstract}

\begin{IEEEkeywords}
 Simultaneous Localization and Mapping (SLAM), Multi-Robot Systems, Optimization, Robust Perception
\end{IEEEkeywords}
\vspace{-0.2cm}

\blfootnote{\hspace{-8.5pt}The authors are with the Robotics Institute, Carnegie Mellon University, Pittsburgh, PA, USA. \texttt{\{danmcgann, kaess\}@cmu.edu}}

\urlstyle{tt} 
\vspace{-20pt}
\section{Introduction}\label{sec:intro}
Collaborative Simultaneous Localization and Mapping (C-SLAM) is a fundamental capability for multi-robot teams~\cite{lajoie_cslam_survey_2022}. A key component of C-SLAM systems is the back-end algorithm that estimates the state of the team from distributed, noisy measurements~\cite{slam_survey_leonard_2016}. However, existing back-ends struggle with the practical conditions experienced by multi-robot teams in the real world. During field deployments (e.g. search and rescue, forestry inspection, or scientific exploration~\cite{multirobot_sar_drew_2022, yuan_fire_survey_2015, cadre_mission_website}), multi-robot teams cannot assume the presence of communication infrastructure, as existing infrastructure may have been damaged (e.g. disaster response) or may not exist (e.g. remote forest, deep sea, lunar). Rather, teams can only assume access to an ad-hoc network that permits unreliable, sparse communication. Additionally, these teams cannot assume that their sensor measurements are perfect. Instead, teams must accept that due to perceptual aliasing,\footnote{Perceptual Aliasing -- The phenomenon where spatially disparate locations display a similar ``perceptual'' (i.e. visual or geometric) appearance.} front-end processes (e.g. loop-closure detection) will produce erroneous outlier measurements. Despite these challenges, multi-robot teams need accurate and timely estimates to support downstream tasks like planning and navigation. Existing C-SLAM back-end algorithms struggle to operate under these conditions, either requiring reliable network connectivity, needing impractical computation time, or failing to provide accurate, robust results. 

In this paper, we present Consensus Alternating Direction Method of Multipliers (C-ADMM) as a framework to design C-SLAM back-ends that can handle these challenging conditions and collaboratively produce high-quality state estimates for multi-robot teams operating in the real world. We specifically propose \textbf{R}obust \textbf{I}ncremental \textbf{M}anifold \textbf{E}dge-based \textbf{S}eparable \textbf{A}DMM (riMESA) -- a robust, incremental, and distributed C-SLAM back-end designed to meet the challenges of real-world operations that outperforms the accuracy of prior works DDF-SAM2~\cite{ddfsam2_cunningham_2013} and DLGBP~\cite{robot_web_journal_murai_2024} by a factor of \textbf{>7x} on real-world C-SLAM problems (Fig.~\ref{fig:rimesa-header-figure}).

The rest of the paper is structured as follows. First, we concretely define the C-SLAM problem that multi-robot teams contend with during real-world deployments and the communications conditions under which it must be solved. Next, we provide an overview of prior works and discuss their ability to solve our target problem. We then introduce C-ADMM, discuss its application to C-SLAM problems, and outline the riMESA algorithm. We conclude with a rigorous evaluation of riMESA on synthetic and real-world data across a variety of C-SLAM scenarios and communication conditions. To aid future research, we make the riMESA code open-source.\footnote{\url{https://github.com/rpl-cmu/rimesa}}

\section{Problem Definition}\label{sec:problem-definition}
We seek to solve the generic C-SLAM problem in which a team of robots $\Rc$ estimates state variables $\Theta$ using noisy measurements $\Mc$. In this multi-robot case, each robot takes a subset of the measurements and observes a subset of the total variables. Importantly, some measurements taken by robots will be inter-robot measurements. These measurements may come from direct observation of other robots, joint observations of environment landmarks, or from a distributed loop-closure system~\cite{distrbuted_loopclosure_tian_2020}. Inter-robot measurements enable the multi-robot team to collaborate. This can improve their individual state estimates and can ensure all robots' solutions remain in a consistent global frame. Due to inter-robot measurements, multiple robots may observe the same variable. Therefore, let $\Theta_i \subset \Theta$ denote the non-disjoint subset of variables observed by robot $i\in \Rc$. Additionally, let $\Mc_i \subset \Mc$ denote the robot's disjoint subset of measurements. 

\begin{remark}[Generic C-SLAM vs. PGO] \label{remark:generic-cslam-vs-pgo}
It is common for C-SLAM back-ends to solve only Pose-Graph Optimization (PGO), a subset of C-SLAM in which all variables are poses $\in$ SE$(N)$ and all measurements are relative poses~\cite{dgs_choudhary_2017, geod_cristofalo_2020, asapp_tian_2020, dc2pgo_tian_2021, maj_min_iros_fan_2020, maj_min_journal_fan_2024}. PGO is useful in many applications and provides structure to exploit in algorithm design. However, it limits map representations (e.g. landmarks) as well as measurement types (e.g. bearing and range), which are highly useful in specific domains (e.g. underwater~\cite{auv_nav_review_paull_2014} and space~\cite{boroson_puffer_range_pgo_2020}). To ensure our method generalizes, we focus on the generic C-SLAM problem.
\end{remark}

\subsection{Batch C-SLAM}\label{sec:problem-definition:batch-cslam}
The de-facto standard for SLAM is to formulate the problem as a factor-graph and solve for the state variables via Maximum A-Posteriori (MAP) inference~\cite{slam_survey_leonard_2016}:
\begin{equation}\label{eq:cslam-as-map}
    \Theta_{MAP} = \argmax_{\Theta \in \Omega} P(\Theta) \prod_{i\in \Rc} P(\Mc_i | \Theta_i)
\end{equation}
where $\Omega$ is the product manifold constructed by the manifold of each variable in $\Theta$. When we assume that each measurement $m\in \Mc$ is affected by zero-mean Gaussian noise with covariance $\Sigma_m$ this problem can be solved by Nonlinear Least Squares (NLS) optimization~\cite{factor_graphs_for_robot_perception}:
\begin{equation}\label{eq:cslam-as-nls}
    \Theta_{MAP} = \argmin_{\Theta \in \Omega} \sum_{i\in\Rc}\sum_{m\in \Mc_i} \norm{h_m(\Theta_i) - m}_{\Sigma_m}^2
\end{equation}
where $h_m(\Theta_i)$ is the measurement prediction function that computes the expected measurement from the state estimate. 

\subsection{Robust Incremental C-SLAM}\label{sec:problem-definition:robust-incremental-cslam}
There are two key challenges when utilizing the C-SLAM problem formulation for real-world deployments. Firstly, we are not interested in just solving this optimization problem once. Rather, for practical robotic applications, we need to solve this problem at every timestep, with the advantage that we have access to the previous solution. Secondly, this NLS formulation is inherently sensitive to measurements that do not exhibit Gaussian noise (i.e. outliers). Due to imperfect processes used to derive our measurements $\Mc$, such outliers are common~\cite{slam_survey_leonard_2016}. Moreover, due to perceptual aliasing, we expect that outliers caused by incorrect data association are inevitable given sufficient operation time. This inevitability, combined with the impact that outliers have on NLS based C-SLAM solvers, necessitates that we extend our algorithms to handle such outliers. Combining these challenges, we define the robust incremental C-SLAM problem:
\begin{equation}\label{eq:robust-incremental-cslam-as-nls}
\begin{aligned}
\Theta^t_{MAP} =& \argmin_{\Theta^t \in \Omega^t} \sum_{i\in\Rc}\sum_{m\in \Ic_i^t} \norm{h_m(\Theta^t_i) - m}_{\Sigma_m}^2 \\
\textrm{given:} \quad & \Theta_{MAP}^{t-1} \\
\end{aligned}
\end{equation}
where $\Ic_i \subseteq \Mc_i$ are the "true" inlier measurements taken by robot $i$ and \eqref{eq:robust-incremental-cslam-as-nls} must be solved online to report results with minimal delay to downstream tasks. For completeness, let $\Oc_i$ represent the set of outliers such that $\Oc_i = \Mc_i \setminus \Ic_i$.

\subsection{Maximum-Consensus}\label{sec:problem-definition:maximum-consensus}
An important nuance in the definition of \eqref{eq:robust-incremental-cslam-as-nls} is that the set of "true" inliers $\Ic$ is unknowable. An intuitive definition of an inlier is any measurement that matches the true state of the world $\Theta^*$ up to a reasonable amount of noise. Given that measurements $m$ are affected by Gaussian noise, we can therefore concretely define an inlier as:
\begin{equation}\label{eq:true_outlier_definition}
    \norm{h_m(\Theta^*) - m }_{\Sigma_m} < \chi_m^2(T)
\end{equation}
where we are using the fact that the residual of the measurement will be distributed according to a  $\chi^2$ distribution with dimensionality matching $m$ and $T$ is a probability threshold (e.g. $T=0.95$). However, the true state of the world $\Theta^*$ is fundamentally unknowable, making the true set of inliers likewise unknowable. A knowable and, therefore, more practical definition we can use to define inliers is that of ``joint-consistency.'' We define a set of measurements $\Jc$ as jointly consistent if there exists some state of the world $\Theta$ such that: 
\begin{equation}\label{eq:joint_consistenct_outlier_definition}
    \norm{h_m(\Theta) - m }_{\Sigma_m} < \chi_m^2(T) \quad\forall\quad m\in\Jc
\end{equation}

Under the assumptions that our robots will generate inlier measurements with non-trivial probability and that outlier measurements are random, it is \textit{likely} that the largest set of jointly consistent measurements corresponds to the true set of inliers. Identifying this largest set of measurements is commonly referred to as the Maximum-Consensus problem, which, despite being an approximation of our goal, is NP-Hard and generally intractable. Therefore, to solve \eqref{eq:robust-incremental-cslam-as-nls} our algorithms will need to find (either explicitly or implicitly) approximate solutions to the Maximum-Consensus problem.

\subsection{C-SLAM Communication}\label{sec:problem-definition:cslam-communication}
In the C-SLAM setting, information is distributed across the team, and robots must communicate to solve the problems discussed above. In real-world applications, the team may not have access to existing communication infrastructure. As such, we can only reliably assume that robots can communicate over an ad-hoc network built from the robots themselves. Due to the scale of the environment, hardware constraints, motion of the robots, and sources of interference, this network will be bandwidth limited, time varying, and frequently disconnected. 

Let us represent the communication network at a specific time $t$ by a sparse undirected graph $\Gc^t = (\Rc, \Ec^t)$ with nodes made up of the robots $\Rc$ and edges $\Ec^t$ defined by the currently available direct connections between robots. This graph is expected to be both sparsely connected at any given timestep and only sparsely connected over time. We assume that robots perform pairwise communications with connected teammates at a relatively low rate (e.g. 0.1 - 10 Hz), where the rate is practically specified by the bandwidth and latency of the underlying communication hardware and protocols. Due to limited connectivity, robots will not be able to maintain synchronized communication, and instead communications will occur asynchronously whenever pairs of robots can connect. We further assume that connections may drop out at any time. Additionally, even if using connection-oriented protocols (i.e. TCP), such dropout may be observable to only one of the communicating robots (i.e. Two-Generals Failure).\footnote{Two-Generals Failure -- Any communication failure where only one party observes that the communication was not successful. Named after the ``Two Generals Problem,'' a thought experiment on unreliable communication~\cite{two_generals_akkoyunlu_1975}.} Finally, we assume that latency may be large relative to the rate of measurements, and communications may need to be performed in parallel with a C-SLAM algorithm.

This ad-hoc, sparse, and unreliable communication model is restrictive and represents a challenging scenario. However, it is representative of what multi-robot teams may encounter during remote field deployments. Additionally, an algorithm that can operate effectively under this restrictive model can also operate in more optimistic scenarios. If the team has use of communication infrastructure, then $\Gc^t$ is simply more densely connected at each timestep. If network bandwidth and robot hardware permits fast communication, then the effective communication rate is simply larger. Finally, if communication links are reliable, then robots simply experience less dropout. Therefore, algorithms that can tolerate this restrictive model will be applicable even when communication is less limited.

\subsection{Problem Scope}\label{sec:problem-definition:proble-scope}
To support real-world deployments of multi-robot teams, we require algorithms that can solve the robust incremental C-SLAM problem \eqref{eq:robust-incremental-cslam-as-nls} even when robots have access to only ad-hoc, sparse, and unreliable communication (Sec.~\ref{sec:problem-definition:cslam-communication}).

\section{Related Work}\label{sec:related-work}
Prior works have not provided an effective solution to the incremental, robust C-SLAM problem with ad-hoc, sparse, and unreliable communication. To address this problem, prior works have broadly sought to combine the independent lines of research on C-SLAM optimization and robust optimization. For both lines of research, there are a number of approaches, which we discuss briefly below. We then discuss the variety of works that have looked to combine these methods to solve our target problem.

\subsection{C-SLAM Optimization}\label{sec:related-work:cslam-opt}
Prior researchers have broadly focused on three classes of C-SLAM optimization architectures.

\subsubsection{Centralized C-SLAM Optimization} \label{sec:related-work:cslam-opt:centralized}
The conceptually simplest approach to the C-SLAM task is to use a centralized back-end. In these methods, robots communicate their measurements to a central server that performs all computation and sends solutions back to the team. Centralized methods have been demonstrated successfully under a variety of conditions~\cite{decent_loc_bailey_2011, mr3dlidar_slam_dube_2017, multi_uav_slam_schmuck_2017, cvi_slam_karrer_2018, ccm_slam_schmuck_2019, covins_schmuck_2021, dist_client_server_slam_zhang_2021, lamp2_chang_2022, cvids_zhang_2022, hydramulti_chang_2023, maplab_cramariuc_2023, rcvislam_pan_2024}. However, they require that $\Gc^t$ is connected at each timestep and can support significant network traffic to return solutions to all team members, making them inapplicable to scenarios with ad-hoc, sparse, and unreliable communications. Additionally, a central server introduces a single point of failure into the system, making these systems brittle. Finally, as these methods combine all measurements from all the robots into a joint optimization problem, centralization struggles to scale to large teams or long-term operation~\cite{subt_lessons_2023}.

\subsubsection{Decentralized C-SLAM Optimization}\label{sec:related-work:cslam-opt:decentralized}
Similar to centralized algorithms are decentralized back-ends. In these approaches each robot holds a copy of the global problem and independently computes the solution~\cite{decent_auv_paull_2014, mr6d_slam_schuster_2015, mr6d_slam_journal_schuster_2019, dense_mr_slam_dubois_2020, disco_slam_haung_2022, swarm_slam_lajoie_2024, slideslam_liu_2025}. While decentralization removes any single point of failure from the system and permits robots to continue operation in the face of sparse communication, these methods require even greater bandwidth than centralized methods to send all measurements to every other robot. Additionally, decentralization necessitates that each robot performs expensive and redundant computation. These issues with bandwidth and computation can be mitigated, but not removed, by sparsifying the global problem~\cite{mr6d_slam_schuster_2015} and sharing computation within robot clusters~\cite{swarm_slam_lajoie_2024}. As such, decentralized methods will, like centralized approaches, struggle to scale to large multi-robot teams and long-term operation.

\subsubsection{Distributed C-SLAM Optimization}\label{sec:related-work:cslam-opt:distributed}
The final class of C-SLAM back-ends are distributed algorithms. Rather than aggregating the global problem, distributed methods allow agents to solve only local sub-problems and rely on information passed during communication to converge local solutions to the global optimum. Distributed algorithms hold the greatest potential for C-SLAM, as local sub-problems can be solved efficiently, algorithms can be designed to tolerate sparse, unreliable communication, and these algorithms can scale to even very large multi-robot teams. However, existing works have not yet taken full advantage of this potential.

Due to the diversity of distributed optimization techniques~\cite{distsurvey1_shorinwa_2024, distsurvey2_shorinwa_2024}, a wide variety of distributed solvers have been proposed. Targeting the batch C-SLAM problem \eqref{eq:cslam-as-nls}, prior works have proposed algorithms based on Multi-Block ADMM (MB-ADMM~\cite{admmslam_choudhary_2015}), Distributed Gauss-Seidel (DGS~\cite{dgs_choudhary_2017}), Distributed Gradient-Descent (GeoD~\cite{geod_cristofalo_2020}), Majorization-Minimization (MM-PGO~\cite{maj_min_iros_fan_2020, maj_min_journal_fan_2024}), and Distributed Riemannian Gradient Descent (DC2-PGO~\cite{dc2pgo_tian_2021}, ASAPP~\cite{asapp_tian_2020}). As batch algorithms, however, these methods are all limited in their ability to address incremental problems. Naively, batch methods could be applied to the incremental case by solving each timestep as a batch problem. However, these methods all require hundreds to thousands of iterations to converge, significantly exceeding the communication and computation availability for any given timestep at which we need a solution. Further, while some of these methods can tolerate asynchronous communication and thus communication dropout between algorithm iterations, they would all require a connected network during the current timestep, which is not guaranteed in real-world scenarios. 

Some prior works have proposed distributed algorithms specifically targeting incremental C-SLAM, namely, Distributed Data Fusion Smoothing and Mapping (DDF-SAM~\cite{ddfsam_cunningham_2010} and DDF-SAM2~\cite{ddfsam2_cunningham_2013}) and Distributed Loopy Gaussian Belief Propagation (DLGBP~\cite{robot_web_journal_murai_2024}). While these methods can achieve real-time performance, due to their underlying optimization strategies, they struggle to provide accurate and consistent results in many scenarios as illustrated in our experimental results (Sec.~\ref{sec:rimesa-experiments}).

\subsection{Robust Optimization}\label{sec:related-work:robust-opt}
Often viewed through the lens of single-robot SLAM (though applicable to C-SLAM), numerous prior works have proposed methods to address outlier measurements by approximately solving the Maximum-Consensus problem.

\subsubsection{Relaxed Consensus}\label{sec:related-work:robust-opt:relaxed-consensus}
One method to approximately solve the Maximum-Consensus problem is to relax joint-consistency \eqref{eq:joint_consistenct_outlier_definition} to measures that are more efficient to compute, like Pairwise Consistency (PCM~\cite{pcm_mangelson_2018}), Group-$k$ Consistency (G$k$CM~\cite{gkcm_forsgren_2024}), or heuristic consistency (iRRR~\cite{irrr_latif_2012, rrr_irrr_latif_2013}, IPC~\cite{ipc_olivastri_2024}). With these relaxed consistency definitions, the largest set of consistent measurements can be estimated with max-clique graph algorithms or repeated optimizations. Despite these approximations, relaxed consensus methods remain computationally expensive, and their consistency approximations limit their ability to derive quality solutions.

\subsubsection{M-Estimation}\label{sec:related-work:robust-opt:mest}
Another way to relax the Maximum-Consensus problem is via M-Estimation, which wraps measurement errors with a robust kernel $\rho$ to mitigate the effects of outliers on the solution~\cite{conic_zhang_1997, mest_convergence_aftab_2015}. In doing so, these methods implicitly approximate a solution to the Maximum-Consensus problem. M-Estimation methods are trivially compatible with any NLS solver, including existing fast and efficient incremental SLAM solvers like iSAM2~\cite{isam2_kaess_2012}, making them able to support real-time operation. However, depending on the selection of kernel~\cite{conic_zhang_1997, dcs_agarwal_2013, maxmix_olson_2013}, M-Estimation methods construct a cost function that either remains sensitive to outliers or becomes increasingly sensitive to initialization~\cite[A6.8]{multiview_hartley_2003}.

\subsubsection{Variable Augmentation}\label{sec:related-work:robust-opt:var-aug}
Lying somewhere between the relaxed consensus methods and M-Estimation methods, variable augmentation approaches add additional variables to the optimization problem that attempt to classify measurements and down-weight the effect of outliers. The resulting problem is solved by an alternating optimization over the state and augmented variables~\cite{switchable_sunderhauf_2012, dcsam_doherty_2022}. However, depending on the exact formulation used for augmented variables, these methods can either construct, via Black-Rangarajan duality, an algorithm that is equivalent to the use of an M-Estimator~\cite{black_rangarajan_duality_1996, adaptive_chebrolu_2021} or one that is computationally inefficient to solve~\cite{dc_convex_opt_lajoie_2019}.

\subsubsection{Continuation}\label{sec:related-work:robust-opt:continuation}
Currently, the most promising approaches to robust SLAM optimization are continuation-based methods. These methods take the same theoretical approach as M-Estimation but resolve the induced initialization sensitivity using iterative optimization over a continuation of the problem. Graduated Non-Convexity (GNC) was the first continuation method proposed for batch SLAM problems~\cite{gnc_yang_2020}. While GNC can provide accurate results, its repeated batch optimizations make it computationally inefficient. However, Robust Incremental Smoothing and Mapping (riSAM) demonstrated that these same ideas could be incrementalized to achieve robust and real-time SLAM optimization~\cite{mcgann_risam_2023}.

\subsection{Robust C-SLAM Optimization}\label{sec:related-work:robust-cslam-opt}
Numerous prior works have looked to make C-SLAM optimization robust by combining existing back-ends with methods for robust optimization. In-turn, these approaches inherit the capabilities and the drawbacks of their components. 

Numerous centralized and decentralized methods have proposed incorporating existing robust optimization methods. Schuster et al.~\cite{mr6d_slam_journal_schuster_2019}, CVI-SLAM~\cite{cvi_slam_karrer_2018}, COVINS~\cite{covins_schmuck_2021}, and RCVI-SLAM~\cite{rcvislam_pan_2024} propose utilizing M-Estimators in their optimization, in-turn inheriting their efficiency but also their induced sensitivity to initialization. Similarly, MAPLAB 2.0~\cite{maplab_cramariuc_2023} proposed the use of Switchable-Constraints, a variable augmentation method that will produce near-equivalent results to M-Estimators. LAMP 2.0~\cite{lamp2_chang_2022}, Hydra-Multi~\cite{hydramulti_chang_2023}, and Swarm-SLAM~\cite{swarm_slam_lajoie_2024} proposed the use of batch GNC to perform their optimization robustly, in-turn inheriting its accuracy but also its inefficiency. CVIDS~\cite{cvids_zhang_2022} and DiSCO-SLAM~\cite{disco_slam_haung_2022} proposed using PCM to filter outliers, in-turn inheriting both its inefficiency and potentially poor performance. In addition to the downsides listed above, all of these methods additionally inherit the communication requirements of centralized and decentralized architectures, making their application to real-world deployments difficult and application to our target problem (Sec.~\ref{sec:problem-definition:proble-scope}) all but infeasible.

Distributed systems have taken similar approaches. DOOR-SLAM~\cite{doorslam_lajoie_2020} proposed using DGS for distributed optimization combined with PCM for outlier-rejection. In doing so, it inherits the computational inefficiency and limited accuracy of DGS as well as the computational inefficiency and limited accuracy of PCM. The same composition and drawbacks can be found in DCL-SLAM~\cite{dcl_slam_zhong_2023}, and a similar composition can be found in $D^2$SLAM~\cite{d2slam_xu_2024}. Kimera-Multi~\cite{kimeramulti_journal_tian_2022} composes the distributed back-end ASAPP with a GNC-based robust estimation approach. Both methods are computationally complex, and composing them multiplicatively increases these costs. Thus, while it can provide accurate results, the method is nearly impossible to apply to the real-time scenarios~\cite{kimeramulti_lessons_tian_2023}. Intentionally targeting real-time performance, Murai et al. proposed combining the use of standard M-Estimators with their proposed DLGBP optimization method~\cite{robot_web_journal_murai_2024}. This combination allows for very efficient incremental solving, as DLGBP can be made efficient and the addition of M-Estimators has little impact on computational performance. However, this also combines DLGBP's poor convergence with a sensitivity to initialization induced by the use of M-Estimators which results in generally poor performance across problem scenarios.

\subsection{Related Work Summary}
Evident from this review, while some prior works can support multi-robot teams in limited conditions, we lack back-end algorithms that can generalize to provide solutions to the incremental, robust C-SLAM problem with ad-hoc, sparse, and unreliable communication. In this work we argue for an alternative approach to the C-SLAM back-end -- distributed consensus ADMM. We outline how C-ADMM can be applied to distributed C-SLAM optimization and, importantly, how it can be extended to effectively address the incremental, robust C-SLAM problem via our algorithm -- riMESA.

\section{Consensus ADMM for Collaborative SLAM}\label{sec:cadmm}

In this section we begin with an overview of C-ADMM theory and then provide an overview of our work that proposed to apply C-ADMM to the C-SLAM problem~\cite{mesa_mcgann_2024}.

\subsection{Standard C-ADMM}\label{sec:cadmm:standard-cadmm}
We begin by reviewing C-ADMM~\cite{cadmm_mateos_2010}, which has long been a popular, fully distributed method for solving consensus optimization problems of the form:
\begin{equation}\label{eq:distributed-consensus-optimization-problem}
\begin{aligned}
    \argmin_{\bar{x}\in\Rb^{|\Rc|n}} \quad & \sum_{i\in \Rc} f_i(x_i) \\
    \textrm{s.t.} \quad & x_i = x_j ~\forall~ (i,j) \in \Ec\\
\end{aligned}
\end{equation}
where $\Rc$ is the set of all agents, $f_i$ is the local objective of agent $i$, $x_i \in  \Rb^n$ is the copy of decision variables held by agent $i$, and $\bar{x}\in \Rb^{|\Rc|n}$ is the concatenation of all agents' local decision variables. Agents communicate over an undirected network made up of the agents~$\Rc$ and communication links~$\Ec$. The constraints force neighboring agents to agree on a single solution. By induction, this forces all agents to converge to a single joint solution to the problem.

Applying standard ADMM to problem \eqref{eq:distributed-consensus-optimization-problem} produces an algorithm that requires centralized updates~\cite{admm_boyd_2010}. To produce a fully distributed algorithm, C-ADMM augments problem \eqref{eq:distributed-consensus-optimization-problem} with additional variables. For each edge in the communication network $(i,j) \in \Ec$, C-ADMM introduces two new variables $z_{(i,j)}$ and $z_{(j,i)}$. The variable $z_{(i,j)}$ is held by agent $i$ and can be interpreted as agent $i$'s estimate of agent $j$'s local solution. We will refer to these variables as "edge variables" and use them to rewrite \eqref{eq:distributed-consensus-optimization-problem} as:
\begin{equation}\label{eq:consensus-admm-problem}
\begin{aligned}
    \argmin_{\bar{x}\in\Rb^{|\Rc|n},~\bar{z} \in \Zc} \quad & \sum_{i\in \Rc} f_i(x_i) \\
    \textrm{s.t.} 
        \quad & x_i = z_{(i,j)},~x_j = z_{(j,i)} ~\forall~ (i,j) \in \Ec
\end{aligned}
\end{equation}
where $\bar{z}$ is the concatenation of all $z_{(i,j)}$ and $\Zc$ is the space of $\Rb^{2|\Ec|n}$ which is implicitly constrained such that $z_{(i,j)} = z_{(j,i)}$. 

This augmentation increases the number of constraints but does not change their meaning. C-ADMM solves \eqref{eq:consensus-admm-problem} according to the update process:
\begin{align}
    \label{eq:standard-admm-updates-x}
    & \bar{x}^{k+1} = \argmin_{\bar{x}\in\Rb^{|\Rc|n}}\Lc(\bar{x}, \bar{z}^k, \bar{\lambda}^k, \beta^k)\\
    \label{eq:standard-admm-updates-z}
    & \bar{z}^{k+1} = \argmin_{\bar{z} \in \Zc}\Lc(\bar{x}^{k+1}, \bar{z}, \bar{\lambda}^k, \beta^k)\\
    \label{eq:standard-admm-updates-dual}
    & \bar{\lambda}^{k+1} = \bar{\lambda}^{k} + \beta^k (D \bar{x}^{k+1} - \bar{z}^{k+1}) \\
    \label{eq:standard-admm-updates-beta}
    & \beta^{k+1} = \alpha \beta^k
\end{align}
where $\bar{\lambda}$ is the concatenation of all dual variables (one corresponding to each constraint), $D\in \Rb^{(2|\Ec|n)\times(|\Rc|n)}$ maps each $x_i$ in $\bar{x}$ to all corresponding $z_{(i,j)}$ in $\bar{z}$, $\beta$ is the penalty term, $\alpha$ is a scaling factor hyper-parameter, $k$ is the iteration count, and $\Lc$ is the problem's Augmented Lagrangian:
\begin{multline}\label{eq:standard-admm-lagrangian}
    \sum_{i\in \Rc} f_i(x_i) + \sum_{j \in \mathcal{N}_i} \inner{\lambda_{(i,j)}}{x_i - z_{(i,j)}} + \frac{\beta}{2}\norm{x_i - z_{(i,j)}}^2
\end{multline}
where $\mathcal{N}_i$ are the neighbors of agent $i$. Solving these iterates can be fully distributed, as \eqref{eq:standard-admm-updates-x} can be solved by each agent minimizing its local Augmented Lagrangian independently. After this minimization, C-ADMM stipulates that all agents communicate their results to all neighbors to allow each agent to independently solve \eqref{eq:standard-admm-updates-z}, \eqref{eq:standard-admm-updates-dual}, and \eqref{eq:standard-admm-updates-beta}. When all $f_i$ are strongly convex, C-ADMM converges linearly~\cite{cadmm_lin_converge_shi_2014}.

Within existing literature, there are two extensions to the base C-ADMM algorithm that will be relevant to its application to C-SLAM problems. Firstly, standard C-ADMM assumes that each agent's objective relies on all optimization parameters. However, there is a large class of problems where each agent's objective relies on only a subset of the variables. We refer to such problems as "separable" consensus optimization problems, a reference to Separable Optimization Variable ADMM (SOVA)~\cite{sova_shorinwa_2020}. SOVA proposed a simple but effective extension in which agents hold only the variables required by their local objective and we impose constraints on only the shared variables. Separable problems take the form:
\begin{equation}\label{eq:sova-optimization-problem}
\begin{aligned}
    \argmin_{\{x_0 \in\Rb^{n_0},~...,~x_r \in\Rb^{n_r} \}} \quad & \sum_{i\in \Rc} f_i(x_i) \\
    \textrm{s.t.} \quad & A_{(i,j)} x_i = B_{(i,j)} x_j ~\forall~ (i,j) \in \Ec\\
\end{aligned}
\end{equation}
where $A_{(i,j)}\in \Rb^{n_{(i,j)} \times n_{i}}$ and $B_{(i,j)}\in \Rb^{n_{(i,j)} \times n_{j}}$ map the variables shared between agents $i$ and $j$ into a common space.

Secondly, C-ADMM assumes an algorithmic structure in which, at each iteration, all agents communicate with all neighbors. However, C-ADMM has been proven to converge for significantly less restrictive communication models. Edge-based C-ADMM assumes that at each iteration of the algorithm, only a subset of communication edges are active, and each agent, therefore, communicates with only some of its neighbors~\cite{thesis_wei_2014}. Without loss of generality, this model can be simplified to assume that at each iteration only two agents communicate with each other. Under this edge-based model, when the problem is convex and each edge is active with positive probability, C-ADMM is proven to converge with rate $O(1/k)$ where $k$ is the number of iterations~\cite{o1k_convergence_wei_2013}.

\subsection{C-ADMM for Batch C-SLAM}\label{sec:cadmm:mesa}
Next, we show how the C-SLAM problem \eqref{eq:cslam-as-nls} can be transformed into an instance of a Separable Consensus ADMM problem with on-manifold decision variables.\footnote{This section is based largely on our prior work~\cite{mesa_mcgann_2024}, though it uses additional insights from follow-up work~\cite{imesa_mcgann_2024} and further experience.}

Starting from the batch C-SLAM problem \eqref{eq:cslam-as-nls} we can apply a distributed lens knowing that measurements $\Mc$ and, in-turn the state variables $\Theta$ are distributed across the robot team. First, let $S_{(i,j)} \triangleq \Theta_i \cap \Theta_j$ represent the variables shared between robots $i$ and $j$, where for $\theta_{s} \in S_{(i,j)}$ we denote $\theta_{s_i}$ as the copy of that variable owned by robot $i$. From this we can rewrite \eqref{eq:cslam-as-nls} to reflect this distribution of information as:
\begin{equation} \label{eq:cslam-as-distributed-nls}
\begin{aligned}
    \Theta_{MAP} = \argmin_{\Theta_i \in \Omega_i \forall i\in\Rc}\quad& 
        \sum_{i\in \Rc} \sum_{m\in \Mc_i} \norm{h_m(\Theta_i) - m}_{\Sigma_m}^2 \\
        \textrm{s.t.} \quad & q_{s}(\theta_{s_i}, \theta_{s_j}) = 0 \\
                            &~\forall~ \theta_s \in S_{(i,j)} ~\forall~(i,j) \in \Ec\\
\end{aligned}
\end{equation}
where $q_{s}$ compares the equality appropriately for the manifold to which $\theta_{s}$ belongs and returns $0 \in \Rb^d$ if and only if $\theta_{s_i}$ and $\theta_{s_j}$ are equal and a ``distance'' between the variables if they are not equal. The generic function $q_{s}$ is used as there exist many ways to compare equality of on-manifold objects.

From \eqref{eq:cslam-as-distributed-nls} we can see that the general C-SLAM problem is an instance of a separable optimization problem \eqref{eq:sova-optimization-problem} as each robot's cost function affects only a subset of the global variables and robots share sparse sets of variables. However, unlike \eqref{eq:sova-optimization-problem} and all standard ADMM formulations, C-SLAM problems typically include on-manifold decision variables, which are handled with generic constraint functions $q_s$.

To permit a fully distributed algorithm, we can take the same approach as standard C-ADMM -- to augment the problem with edge variables. Specifically, for every shared variable $\theta_s$ shared along edge $(i,j)$, we introduce edge variables $z_{(i,j)_s}$ and $z_{(j,i)_s}$. The addition of edge variables allows us to rewrite the constraints of the problem as:
\begin{equation} \label{eq:cslam-as-ebadmm}
\begin{aligned}
    \Theta_{MAP} = \argmin_{\Theta_i \in \Omega_i \forall i\in\Rc,~Z \in \Zc}\quad& 
        \sum_{i\in \Rc} \sum_{m\in \Mc_i} \norm{h_m(\Theta_i) - m}_{\Sigma_m}^2 \\
        \textrm{s.t.} \quad & q_{s}(\theta_{s_i}, z_{(i,j)_s}) = 0 \\
                            & q_{s}(\theta_{s_j}, z_{(j,i)_s}) = 0 \\
                            &~\forall~ \theta_s \in S_{(i,j)} ~\forall~(i,j) \in \Ec\\
\end{aligned}
\end{equation}
where $Z$ is the set of all $z_{(i,j)_s}$ and $\Zc$ is the appropriate product manifold further constrained such that $z_{(i,j)_s} = z_{(j,i)_s}$. We note here that the representation used for edge variables will depend in part on the representation of its corresponding state variable $\theta_s$ and the selection of constraint function $q_s$. Next we use (\ref{eq:standard-admm-updates-x},~\ref{eq:standard-admm-updates-z},~\ref{eq:standard-admm-updates-dual},~\ref{eq:standard-admm-updates-beta},~\ref{eq:standard-admm-lagrangian}) to derive on-manifold, separable, C-ADMM iterates required to solve \eqref{eq:cslam-as-ebadmm}. 
\begin{align}
    & \label{eq:mesa-updates-x}
    \begin{aligned}
    \Theta^{k+1}_{i} =& \argmin_{\Theta_i \in \Omega_i} \quad 
          \sum_{m\in \Mc_i} \norm{h_m (\Theta^k_i) - m}_{\Sigma_m}^2 \\
          &+ \sum_{j \in \mathcal{N}_i} \sum_{s\in S_{(i, j)}} \inner{\lambda^k_{(i,j)_s}}{q_{s}\left(\theta_{s_i}, z^{k}_{(i,j)_s}\right)} \\
          &+ \sum_{j \in \mathcal{N}_i} \sum_{s\in S_{(i, j)}} \frac{\beta^k}{2} \norm{q_{s}\left(\theta_{s_i}, z^{k}_{(i,j)_s}\right)}^2\\
    \end{aligned}\\
    & \label{eq:mesa-updates-z}
    \begin{aligned}
         {z^{k+1}_{(i, j)_s}}& = 
          \argmin_{z_s \in \Zc_s}\\ 
          &\inner{\lambda^k_{(i,j)_s}}{q_{s}\left(\theta^{k+1}_{s_i}, z_s\right)} + \frac{\beta^k}{2} \norm{q_{s}\left(\theta^{k+1}_{s_i}, z_s\right)}^2 \\
         +&\inner{\lambda^k_{(j,i)_s}}{q_{s}\left(\theta^{k+1}_{s_j}, z_s\right)} + \frac{\beta^k}{2} \norm{q_{s}\left(\theta^{k+1}_{s_j}, z_s\right)}^2 \\
    \end{aligned}\\
    & \label{eq:mesa-updates-dual}
    \lambda_{(i,j)_s}^{k+1} = \lambda_{(i,j)_s}^{k} + \beta^k \left(  q_{s}\left(\theta^{k+1}_{s_i}, z^{k+1}_{(i,j)_s}\right) \right) \\
    \label{eq:mesa-updates-beta}
    & \beta^{k+1} = \alpha \beta^k
\end{align}%
where a dual variable $\lambda_{(i,j)_s}$ is introduced for each constraint and the iterates are written for individual variables where they can be solved for independently. However, to actually apply these iterates to real problems, we need methods to solve the various Augmented Lagrangian optimization problems.

\subsubsection{State Variable Update} \label{sec:cadmm:mesa:updates-x}
We begin by looking at the state variable update \eqref{eq:mesa-updates-x}. To solve this update, we can leverage existing NLS solvers by transforming the problem slightly, combining the dual and penalty terms of the Augmented Lagrangian into a ``Biased Prior'' (Sec.~\ref{sec:cadmm:mesa:biased-priors}). This reformulation allows us to compute solutions to \eqref{eq:mesa-updates-x} using existing sparse NLS solvers like \texttt{gtsam}, \texttt{g2o}, or \texttt{ceres}~\cite{dellaert_gtsam_tech_report_2012, g2o_kummerle_2011, ceres_agarwal_2022}.

\subsubsection{Edge Variable Update} \label{sec:cadmm:mesa:updates-y}
We next look at the edge-variable updates \eqref{eq:mesa-updates-z}. Using a similar reformulation used for the $\Theta$ update, we could construct and solve an NLS optimization problem for each $z^{k+1}_{(i, j)_s}$. However, as robots may share large numbers of variables, to improve efficiency, we would prefer that \eqref{eq:mesa-updates-z} be solved in closed form. As we will later show, depending on the selection of constraint function and how we choose to represent our edge variables (Sec.~\ref{sec:cadmm:mesa:constraint-functions}), a closed-form solution may exist or may be approximated. 

\subsubsection{Dual Updates} \label{sec:cadmm:mesa:updates-dual}
Computing updates to dual variables is straightforward according to \eqref{eq:mesa-updates-dual} once a constraint function and edge variable representation have been selected.  

\subsubsection{Penalty Updates} \label{sec:cadmm:mesa:updates-beta}
For applications that require very precise solutions, the penalty parameter $\beta$ can be scaled over time using parameter $\alpha$ using a straightforward update \eqref{eq:mesa-updates-beta}.

\subsubsection{Biased Priors} \label{sec:cadmm:mesa:biased-priors}
Originally identified by Choudhary et al.~\cite{admmslam_choudhary_2015}, the Augmented Lagrangian in \eqref{eq:mesa-updates-x} can be represented as an NLS problem by combining the dual and penalty terms as ``Biased Priors''. Specifically, the identity that $\argmin_a \inner{b}{a} +(\beta/2)\norm{a}^2 = \argmin_a (\beta/2)\norm{a + b/\beta}^2$ when $b$ is constant is used to rewrite the terms as \eqref{eq:biased-prior}.
\begin{equation}
\label{eq:biased-prior}
\frac{\beta_{(i,j)}}{2} \norm{q_{s}\left(\theta_{s}, z_{(i,j)_s}\right) + \frac{\lambda_{(i,j)_s}}{\beta_{(i,j)}}}^2 
\end{equation}

Representing consensus constraints, biased priors are effectively the mechanism by which all robots' local solutions are constrained to be equal for any shared variables -- thus ensuring a single consistent state estimate for the team.

However, there is a practical challenge in applying biased priors. In most C-SLAM problems we are optimizing over poses that live on the $\mathrm{SE}(N)$ manifold. These poses are challenging as the translation ($t$) and rotation ($r$) components live in different domains with different scales $t \in [-D, D]$ and $r\in (-\pi, \pi]$ where $D$ is the size of the team's operational area. However, a biased prior, as written in \eqref{eq:biased-prior}, treats these components uniformly. The practical effect is that these biased priors under-constrain a pose's rotational components, which in-turn can cause stability and convergence issues when solving for $\Theta_i^{k+1}$. To improve stability, it is best to account for the different scales of these domains by adding a noise model $\Sigma_s$. Specifically, we propose using a noise model $\Sigma_s$ with $\sigma_r = 0.1$ and $\sigma_t = 1$. We refer to biased priors with $\Sigma_s\neq I$ as ``Weighted Biased Priors'' for clarity.

\subsubsection{Constraint Functions}\label{sec:cadmm:mesa:constraint-functions}
In the standard C-ADMM literature (Sec.~\ref{sec:cadmm:standard-cadmm}), all algorithms assume vector-valued variables and thus that all constraints are linear ($q(a,b) \triangleq a-b$), which permits a closed-form solution to \eqref{eq:mesa-updates-z} of $z_{(i,j)_s}^{k+1} = \frac{1}{2}(\theta^{k+1}_{s_i} + \theta^{k+1}_{s_j})$. For vector-valued variables in C-SLAM problems (e.g. landmarks), this approach should be taken. 

However, for the $\mathrm{SE}(N)$ variables typically found in C-SLAM problems, the equality comparison is less straightforward. In Tab.~\ref{tab:constraint_functions} we outline four possible constraint functions and for each describe the closed-form solution to \eqref{eq:mesa-updates-z} or an approximate solution if no closed-form solution exists.

\begin{table}[ht]
\caption{Constraint functions for $\mathrm{SE}(N)$ objects and their corresponding closed-form solutions to \eqref{eq:mesa-updates-z}. Where SPLIT interpolates the translation component linearly and the rotation component spherically~\cite{interpolation_survey_haarbach_2018}, Vec returns a vector of the objects non-constant matrix elements~\cite{madamm_kovnatsky_2016}, and $p$ is the dimension of the tangent space for SE$(N)$. Manifold object notation is derived from the work of Solà et al. \cite{microlie_sola_2021}. The geodesic $z$ update is approximated by the case where $\lambda_{(i,j)_s} = \lambda_{(j,i)_s} = \mathbf{0}$.}
\centering
\scriptsize
\begin{tabular}{|Sl|Sc|Sc|Sc|}
\hline
Function  & $z\in$            & $q(\theta, z)$                                      & $z^{k+1}_{(i,j)_s}$ \\ \hline\hline
Geodesic  & $\mathrm{SE}(N)$  & $\logmap{z^{-1} \circ \theta}$                      & $\mathrm{SPLIT}\left(\theta_{s_i}, \theta_{s_j}, 0.5\right)$ \\  \hline
Apx.-Geo. & $\Rb^{p}$         & $\logmap{\theta} - z$                               & $\frac{1}{2}\left(\logmap{\theta_{s_i}} + \logmap{\theta_{s_j}}\right)$  \\ \hline
Split     & $\mathrm{SE}(N)$  & $\mat{\logmap{R_z^{-1}R_\theta} \\ t_\theta - t_z}$ & $\mathrm{SPLIT}\left(\theta_{s_i}, \theta_{s_j}, 0.5\right)$ \\ \hline
Chordal   & $\Rb^{N^2+N}$     & $\mathrm{Vec}(\theta) - z$                          & $\frac{1}{2}\left( \mathrm{Vec}\left(\theta_{s_i}\right) + \mathrm{Vec}\left(\theta_{s_j}\right)\right)$ \\ \hline
\end{tabular}
\label{tab:constraint_functions}
\vspace{-0.1cm}
\end{table}

While, on paper, all the formulations in Tab.~\ref{tab:constraint_functions} appear reasonable, our prior investigations~\cite{mesa_mcgann_2024} reveal that the geodesic formulation provides the best performance when applied to C-SLAM problems. This finding agrees with general findings from the SLAM community, which has broadly converged to use $\mathrm{SE}(N)$ representations and geodesic error for optimization over robot poses.

\subsubsection{Asynchronous Communication}\label{sec:cadmm:mesa:asynchronous-communication}
To compute the edge and dual variable updates, robots must communicate their current state estimates $\Theta_i$ for any shared variables. Many batch distributed C-SLAM optimization algorithms require that such communication occurs synchronously, with each robot communicating in lockstep. However, this C-ADMM-based approach to C-SLAM permits an edge-based communication model (Sec.~\ref{sec:cadmm:standard-cadmm}) in which agents perform asynchronous pairwise communications. To support this model, we do need to add unique penalty terms $\beta_{(i,j)}, \beta_{(j,i)}$ for each pair of robots, as update \eqref{eq:mesa-updates-beta} may be run at different effective rates.

In batch scenarios, the use of an edge-based communication model practically just means less downtime, as robots do not have to wait for rounds of communication to complete and can better utilize communication and computational resources. However, we will see later that the capability to support asynchronous communication is a key advantage of C-ADMM-based approaches when applied in incremental scenarios.

\subsubsection{MESA+}\label{sec:cadmm:mesa:summary}
Putting together the ideas from this section (separable and on-manifold C-ADMM iterates, weighted biased prior reformulation, geodesic constraint selection, and an edge-based asynchronous communication model), we derive the Manifold, Edge-Based, Separable ADMM Plus (MESA+) algorithm (Alg.~\ref{alg:mesa+}), which optimizes the following iterates:
\begin{align}
    & \label{eq:mesa+-updates-x}
    \begin{aligned}
    \Theta^{k+1}_{i}& = \argmin_{\Theta_i \in \Omega_i} \quad 
          \sum_{m\in \Mc_i} \norm{h_m \left(\Theta^k_i\right) - m}_{\Sigma_m}^2 \\
          &+ \sum_{j \in \Rc} \sum_{s\in S_{(i, j)}}
             \frac{\beta^k_{(i,j)}}{2} \norm{\logmap{\theta^k_{s_i} \ominus z^k_{(i,j)_s}}  + \frac{\lambda^k_{(i,j)_s}}{\beta^k_{(i,j)}}}_{\Sigma_s}^2 
    \end{aligned}\\
    & \label{eq:mesa+-updates-z}
    \begin{aligned}
         {z^{k+1}_{(i, j)_s}}& = \mathrm{SPLIT}\left(\theta^{k+1}_{s_i}, \theta^{k+1}_{s_j}, 0.5\right)
    \end{aligned}\\
    & \label{eq:mesa+-updates-dual}
    \lambda_{(i,j)_s}^{k+1} = \lambda_{(i,j)_s}^{k} + \beta_{(i,j)}^k \logmap{\theta^{k+1}_{s_i} \ominus z^{k+1}_{(i,j)_s}} \\
    \label{eq:mesa+-updates-beta}
    & \beta_{(i,j)}^{k+1} = \alpha \beta_{(i,j)}^k
\end{align}

\begin{algorithm}
\scriptsize 
\captionsetup{font=scriptsize} 
\caption{Manifold, Edge-based, Separable, ADMM Plus (MESA+) }\label{alg:mesa+}
    \begin{algorithmic}[1]
    \State \textbf{In:} Robots $\Rc$, communication links $\Ec$, local estimates $\left\{\Theta^0_0, \Theta^0_1, ..., \Theta^0_r\right\}$
    \State \textbf{Out:} Final Variable Estimates $\left\{\Theta^{final}_0, \Theta^{final}_1, ..., \Theta^{final}_r\right\}$
    \State $\lambda_{(i,j)_s}, \lambda_{(j,i)_s} \gets \mathbf{0}~\forall~ s \in S_{(i,j)} ~\forall~ (i,j) \in \Ec$
    \State $z_{(i,j)_s} \gets \theta^0_{s_i} , z_{(j,i)_s} \gets \theta^0_{s_j} ~\forall~ s \in S_{(i,j)} ~\forall~ (i,j) \in \Ec$
    \While{Not Converged}
        \If{Communication available between robot $i$ and robot $j$}
            \State In parallel update $\Theta_i$ and $\Theta_j$ with \eqref{eq:mesa+-updates-x}
            \State Between $i$ and $j$ communicate $\theta_{s_i}, \theta_{s_j} ~\forall~\theta_s \in S_{(i,j)}$
            \State In parallel update $z_{(i,j)_s}, z_{(j,i)_s} ~\forall~\theta_s \in S_{(i,j)}$ with \eqref{eq:mesa+-updates-z}
            \State In parallel update $\lambda_{(i,j)_s}, \lambda_{(j,i)_s} ~\forall~\theta_s \in S_{(i,j)}$ with \eqref{eq:mesa+-updates-dual}
            \State In parallel update $\beta_{(i,j)}^{k+1}$ and $\beta_{(j,i)}^{k+1}$ with \eqref{eq:mesa+-updates-beta}
        \EndIf
    \EndWhile
    \end{algorithmic}
\end{algorithm}

\begin{remark}[MESA vs. MESA+]
    We highlight that MESA+ bears differences from our previously proposed MESA~\cite{mesa_mcgann_2024} (prompting the ``+'' to differentiate). Namely, MESA+ utilizes weighted biased priors, selects the best-performing constraint function, and does not adopt MESA's alternate dual update (\cite[Eq.(18)]{mesa_mcgann_2024}) as our experience shows that it can sometimes decrease numerical stability.
\end{remark}

\section{Robust Incremental MESA}\label{sec:rimesa}

In this section we propose riMESA, an extension of the MESA+ algorithm developed to solve the robust, incremental problem outlined in Sec.~\ref{sec:problem-definition:proble-scope}. As discussed in Sec.~\ref{sec:related-work}, batch C-SLAM back-end algorithms cannot be naively applied to the incremental problem. MESA+, however, provides an algorithmic structure that permits us to design an effective, robust, and incremental version.

\subsection{Incrementalization Overview}\label{sec:rimesa:incrementalization-overview}
An iteration of MESA+ can be broadly broken down into two steps -- a local optimization step \eqref{eq:mesa+-updates-x} and a communication step (\ref{eq:mesa+-updates-z}, \ref{eq:mesa+-updates-dual}, \ref{eq:mesa+-updates-beta}). In the local optimization step, robots compute their local estimate, pushing this solution to equal that of the other robots via weighted biased priors. In the communication step, robots share relevant portions of their estimates and tighten shared variable constraints to provide better consistency. It is over many iterations that shared variable constraints are fully realized and a final consistent solution is reached.

A key idea of riMESA is to amortize this process (i.e. the tightening of constraints) over time as communication is available between robots. This amortization is possible for three reasons. Firstly, and most importantly, intermediate results (i.e. from iterations in which shared variable constraints are not tight) are still useful. Solutions to \eqref{eq:mesa+-updates-x} are effectively lower bounded by that found by a robot operating independently and improved with even partially tight constraints. Secondly, C-ADMM-based algorithms demonstrate fast convergence, and therefore consistency and solution quality improve quickly even when the algorithm is amortized over time. Finally, MESA+ is fully agnostic to the optimization process used to solve \eqref{eq:mesa+-updates-x}. It only requires that we have a solution when communication occurs. This permits us to use existing incremental optimization methods to achieve online performance. 

This idea gives rise to an asynchronous, two-step algorithm consisting of a local update step in which a robot incrementally updates their local solution as new measurements are observed and a communication step in which a pair of robots update their edge and dual variables to tighten shared variable constraints whenever communication is available.

\begin{remark}[Batch C-SLAM Amortization] \label{remark:batch-cslam-amortization}
    This structure of amortization is not compatible with all batch C-SLAM algorithms. Algorithms based on distributed gradient descent~\cite{dgs_choudhary_2017, dc2pgo_tian_2021, asapp_tian_2020, geod_cristofalo_2020} do not provide useful results at intermediate steps of the algorithm. C-ADMM-based algorithms provide a unique basis for an incremental back-end.
\end{remark}

\subsection{Robustness Overview}\label{sec:rimesa:robust-distribued-opt-overview}
Incremental optimization methods that solve \eqref{eq:mesa+-updates-x}, however, remain sensitive to outliers due to their NLS formulation. Due to the success of continuation methods (Sec.~\ref{sec:related-work:robust-cslam-opt}), we next look to address this sensitivity by reformulating the problem through the lens of M-Estimation.

Returning to the distributed C-SLAM optimization problem \eqref{eq:cslam-as-distributed-nls}, with the knowledge that some measurements may be incorrect, we can apply robust kernels $\rho$ to our measurements to downweight the effects of outliers:
\begin{equation} \label{eq:cslam-as-robust-distributed-nls}
\begin{aligned}
    \Theta_{MAP} = \argmin_{\Theta_i \in \Omega_i \forall i\in\Rc}\quad& 
        \sum_{i\in \Rc} \sum_{m\in M_i} \rho\left(\norm{h_m(\Theta_i) - m}^2_{\Sigma_m}\right) \\
        \textrm{s.t.} \quad & q_{s}(\theta_{s_i}, \theta_{s_j}) = 0 \\
                            &~\forall~ \theta_s \in S_{(i,j)} ~\forall~(i,j) \in \Ec\\
\end{aligned}
\end{equation}

From \eqref{eq:cslam-as-robust-distributed-nls}, we can augment the problem with edge-variables and derive the C-ADMM iterates as done in Sec.~\ref{sec:cadmm:mesa}. Doing so results in nearly the exact same iterates shown above (\ref{eq:mesa+-updates-x}, \ref{eq:mesa+-updates-z}, \ref{eq:mesa+-updates-dual}, \ref{eq:mesa+-updates-beta}), with the only difference being in the $\Theta$ update \eqref{eq:mesa+-updates-x} where $\|h_m \left(\Theta^k_i\right) - m\|_{\Sigma_m}^2$ becomes $\rho(\|h_m \left(\Theta^k_i\right) - m\|_{\Sigma_m}^2)$.

Though seemingly a small change to the mathematics, this change highlights an exciting capability of C-ADMM-based optimization for distributed C-SLAM -- that outliers need only be handled locally. To handle outlier measurements, one only needs to perform the local optimization robustly. Of course, to additionally support the proposed incrementalization, we also require that we can perform this robust local optimization efficiently. Towards this, we propose using riSAM~\cite{mcgann_risam_2023}. As a continuation method, riSAM achieves robustness without inducing a sensitivity to initialization, and as an incremental method, riSAM is able to maintain real-time efficiency.

\subsection{Communication Overview}\label{sec:rimesa:communication-overview}
We next turn our attention to the communication step of MESA+ (\ref{eq:mesa+-updates-z}, \ref{eq:mesa+-updates-dual}, \ref{eq:mesa+-updates-beta}). Due to its basis in C-ADMM, MESA+ permits asynchronous communication. This in-turn allows the proposed algorithm riMESA to natively handle sparse asynchronous communication between agents we expect from our communication networks (Sec.~\ref{sec:problem-definition:cslam-communication}). However, there are additional challenges of real-world communication networks that we must address -- latency and communication failures.

Due to network latency, the time to complete a communication between two robots could be longer than the delay between new local measurements with which we need to update our robot's state estimate. Additionally, at any point during this communication, the connection between robots could fail. To ensure that communication/measurement updates do not conflict and that mid-communication failures do not break the algorithm, we adopt a ``Communication Handler.'' 

The handler holds a cached copy of the algorithm's state and operates in a separate thread to manage communication between robots. By operating in a separate thread, the main thread can continue to perform updates as new measurements arrive. Additionally, by operating on only a cached copy of the algorithm's state, no changes affect the main thread until the communication is complete and successful. Thus, if there is a failure mid-communication, the handler can simply be discarded, and the only effect on the algorithm is that the communication opportunity is not utilized. Notably, this structure makes it trivial to support multiple parallel communications.

However, despite this design, Two-Generals failures can still cause issues. In the case of such a failure, only one robot will incorporate information from a communication. This will technically violate the invariant stipulated by our C-ADMM design that $z_{(i,j)_s} = z_{(j,i)_s}$. While we cannot prevent these types of failures~\cite{two_generals_akkoyunlu_1975}, we design our algorithm to ensure that robots are always able to re-synchronize on their next communication.

\subsection{Additional Details}\label{sec:rimesa:additional-details}
Stemming from interactions between incremental distributed operation and robust optimization, there are a few additional challenges that we now address. 

\subsubsection{Robust Weighted Biased Priors}\label{sec:rimesa:additional-details:robust-biased-priors}
By pursuing M-Estimation based approaches, during our local optimization we are simultaneously estimating the variable state as well as attempting to correctly classify the measurement as either an inlier or an outlier. From an information-theoretic perspective, the information that could provide a clear classification for an inter-robot measurement could come from either robot. Importantly, in incremental settings this information can be observed at any time. Consider the case of a misclassified measurement to which one robot later observes clarifying information. Over time, communication between the two robots will allow information to ``flow'' and allow both to correct the misclassification, but this can take a prohibitively long time due to sparse communication between agents. 

To allow robots to more quickly respond to information, we additionally wrap all biased priors in robust kernels. We refer to these as ``Robust Weighted Biased Priors'' (RWBP), and their use results in the state variable update:
\begin{equation} \label{eq:robust-mesa-updates-x-robust-bp}
\begin{aligned}
    \Theta^{k+1}_{i}& = \argmin_{\Theta_i \in \Omega_i} \quad 
    \sum_{m\in \Mc_i} \rho\left(\norm{h_m \left(\Theta^k_i\right) - m}_{\Sigma_m}^2\right) \\
    + \sum_{j \in \Rc} &\sum_{s\in S_{(i, j)}}
     \rho\left(\frac{\beta^k_{(i,j)}}{2} \norm{\logmap{\theta^k_{s_i} \ominus z^k_{(i,j)_s}}  + \frac{\lambda^k_{(i,j)_s}}{\beta^k_{(i,j)}}}_{\Sigma_s}^2\right)
\end{aligned}
\end{equation}

This allows robots to effectively consider the constraints that biased priors represent as ``out-of-date'' in-turn allowing them to better utilize new information to correct misclassifications and maintain higher quality state estimates.

\subsubsection{Dual Variable Decay}\label{sec:rimesa:additional-details:dual-decay}
While the use of robust weighted biased priors solves an important challenge, it also induces another. With robots able to reject constraints, we need to address their corresponding variables $z_{(i,j)_s}$ and $\lambda_{(i,j)_s}$. The edge-variable $z_{(i,j)_s}$ will be updated at the next communication opportunity. Out-of-date dual variables, however, will continue to affect the algorithm as they have been aggregated into the current value $\lambda_{(i,j)_s}$. While dual variables are adaptive in both magnitude and direction as local estimates change, jumps from constraint rejection can leave dual variables with large components in arbitrary dimensions. To tackle this challenge, we introduce a decay rate $\mathfrak{d}=0.9$, which we use to downweight old dual estimates so that recent information dominates the dual's influence. This is added to \eqref{eq:mesa+-updates-dual}, resulting in:
\begin{equation} \label{eq:rimesa-updates-dual-with-decay}
\begin{aligned}
    \lambda_{(i,j)_s}^{k+1} = \mathfrak{d}\lambda_{(i,j)_s}^{k} + \beta^k \left(  q_{s}\left(\theta^{k+1}_{s_i}, z^{k+1}_{(i,j)_s}\right) \right) \\
\end{aligned}
\end{equation}

\subsubsection{Penalty Parameters \& RWBP Initialization}\label{sec:rimesa:additional-details:fixed-penalty-parameters}
In the incremental case, we add new measurements and new RWBPs as they are observed, which results in a handful of challenges.

First is that after a new shared variable is observed via an inter-robot measurement and its corresponding RWBP is constructed, there may be a period of time before the pair of robots communicates. During this time, the inter-robot measurement and corresponding RWBP provide no information, as the robot lacks a valid estimate of $z_{(i,j)_s}$. We refer to RWBPs during this period as ``uninitialized.'' To ensure that uninitialized RWBPs do not affect the solution, we construct the penalty parameter with a very small value ($\beta_{uninit}$). As other RWBPs may have been previously initialized for the pair, this necessitates that we track a unique $\beta_{(i,j)_s}$ for all shared variables $s$. 

When robots successfully communicate for the first time after a new RWBP is constructed and we compute values for its variables ($z_{(i,j)_s}$, $\lambda_{(i,j)_s}$, $\beta_{(i,j)_s}$), we consider the RWBP to be ``initialized.'' As discussed in Sec.~\ref{sec:cadmm:mesa:updates-beta}, after initialization, updating the penalty parameter can be used to tighten the shared variable constraints. However, this can cause challenges when loop-closures are observed or when constraints are rejected, as discussed above. If constraints are enforced with a large penalty parameter, it can prevent a robot's local optimization process from updating its solution with new information. Therefore, we use only dual variable updates to tighten constraints and use a constant penalty parameter value $\beta_{init}$ once an RWBP is initialized. 

\subsubsection{Shared Variable Initialization}\label{sec:rimesa:additional-details:shared-variable-initialization}
When RWBPs are initialized, we also initialize the corresponding shared variable $\theta_{s_i}$. In general, C-ADMM-based algorithms are not overly sensitive to initial values and can, without any loss of generality, use an initial estimate provided locally or use the current solution from whichever robot ``owns'' the shared variable.

However, in the presence of outlier measurements, the choice of initialization injects information into the optimization process. Given that we don't know what information will allow for correct measurement classification, we avoid throwing away information. To do so, we require robots to utilize their local solution along with the inter-robot measurement to initialize any new shared variable. One challenge to this invariant is that some measurements do not observe a variable's entire state. For example, a bearing and range measurement observes position directly but not the orientation of another robot. To account for this challenge, we add a robust initialization scheme for shared variables. Robots will use a local estimate when new shared variables are observed.\footnote{Note that default values (e.g. $\mathbf{0}$) are needed for unobserved components.} When communication occurs and we initialize the corresponding biased prior, robots will overwrite their estimate using the owner's value for any component that is not observed locally. 

\subsection{riMESA Implementation Details}\label{sec:rimesa:implementation-details}
To realize the ideas proposed above, we must handle the practicalities of implementing the algorithm. Firstly, let each robot $i$ maintain the following internal state:
\begin{center}
\scriptsize
\renewcommand*{\arraystretch}{1.2}
\begin{tabular}{m{0.18\linewidth}|m{0.75\linewidth}}
    \texttt{risam} & An instance of an riSAM optimizer~\cite{mcgann_risam_2023}. \\
    \hline
    $\Theta$ & The local state estimate. \\
    \hline
    $S_j ~\forall~j\in\Rc$ & Sets variables shared with other robots. \\ 
    \hline
    $\Yc_j~~\forall~j\in\Rc$ & Variables needing initialization from other robots. \\
    \hline
    $E$ & The set of observed environment variables. \\ 
    \hline
    $\Lambda = \{ \lambda_{(i,j)_s}\}$ & The set of all dual variables.   \\
    \hline
    $Z = \{z_{(i,j)_s}\}$ & The set of all edge variables.\\
    \hline
    $B = \{\beta_{(i,j)_s}\}$ & The set of all penalty parameters.\\
    \hline
    $\Cc$ & Cache of biased priors to be added the optimizer. \\
    \hline
    $\Kc$ & Cache of variables affected by new communication. \\ 
    \hline
    $\Wc$ & Cache of variables marked for re-convexification. \\
    \hline
    $\Qc$ & Mapping from shared variables to the local variables to which they are connected.  \\
    \hline
    $\Ac$ & Mapping from shared variables to their local observability.    \\
\end{tabular}
\end{center}
Secondly, let each robot $i$ be configured with the following hyperparameter values:
\begin{center}
\scriptsize
\renewcommand*{\arraystretch}{1.2}
\begin{tabular}{ l c m{0.65\linewidth} }
    $\beta_{uninit}$ & $1e^{-4}$ & Penalty parameter for uninitialized biased priors. \\
    \hline
    $\beta_{init}$ & 1.0 & Penalty parameter for biased priors.
\end{tabular}
\end{center}
The exact use of these state variables will be illustrated concretely in the following sections.

\subsubsection{Bookkeeping} \label{sec:rimesa:implementation-details:bookkeeping}
Unlike batch settings, in incremental C-SLAM, the underlying problem is changing at every timestep as robots take new measurements. We must, therefore, efficiently track information required to perform local updates, identify shared variables, and communicate information.

To facilitate communication, each robot $i$ tracks $S_j~\forall~j\in\Rc$, which holds the set of variables locally known to be shared between the pair $(i,j)$, and $E$, which holds the set of environment variables observed by robot $i$. When a robot $i$ makes a new inter-robot measurement that observes a shared variable not already in $S_j$ for the respective robot $j$, it is added to $S_j$ and marked for initialization in $\Yc_j$. When a robot observes environmental state (e.g. landmarks), it does not know what other robots with which it shares this state. Therefore, we add environment variables to a separate set $E$ handled specially.\footnote{Environment Variable -- Named to make clear that these variables do not represent the state of any robot but rather represent the state of an element of the local environment.} 

\begin{remark}[Shared Variable Identification]\label{remark:shared_variable_identification}
Identification of shared state from indirect measurements will be unambiguous, as the distributed loop-closure system that derived the measurement will know the exact data used to derive the measurement, from which robot that data was received, and the identifier that robot uses for the data. Identification via direct measurements is more challenging, as identification depends on the measurement front-end. However, in general we propose that they can be identified by timestamp, given that robots' clocks are sufficiently synchronized and that the measurement front-end is able to identify which teammate is being observed. Identification of shared environment variables like landmarks is more difficult. Practically, robots will likely need to compare some form of descriptor to identify any shared state. We assume that $E$ holds both variable identifiers and descriptors. We further assume that comparison between such sets (i.e. $E \cup E'$) evaluates descriptor similarity to find shared environment state.
\end{remark}

For each variable $s$ shared with another robot $j$, each robot $i$ must also maintain a corresponding dual variable $\lambda_{(i,j)_s}$, edge variable $z_{(i,j)_s}$, penalty parameter $\beta_{(i,j)_s}$, and robust weighted biased prior (Sec.~\ref{sec:rimesa:additional-details:robust-biased-priors}). Thus, when observing a new shared variable, a robot must extend $\Lambda$, $Z$, and $B$ as well as construct a corresponding biased prior that references these new values. Additionally, to support robust optimization, we track in $\Qc$ which local variables are connected to any shared variable $s$, and to support robust initialization, we track in $\Ac$ the local observability of all shared variables. All together the bookkeeping process is summarized in Alg.~\ref{alg:rimesa-bookkeeping}.

\begin{algorithm}
\scriptsize 
\captionsetup{font=scriptsize} 
\caption{riMESA: Bookkeep (Local to robot $i$)}\label{alg:rimesa-bookkeeping}
    \begin{algorithmic}[1]
    \State \textbf{In:} New environment variables, $E_{new}$, new shared variables $S_{new}$, their initial estimates $\Phi$, and a flag $L$ indicating local observation
    \State Extend $E$ for $e$ in $E_{new}$
    \For{each variable $s$ in $S_{new}$ shared with robot $j$}
        \State Extend $S_j$ for new variable $s$
        \State Extend $\Yc_j$ for new variable $s$ if owned by $j$
        \State Extend $\Lambda$ for $j,s$ with value $\mathbf{0}$
        \State Extend $Z$ for $j,s$ with value $\Phi_s$
        \State Extend $B$ for $j,s$ with value $\beta_{uninit}$
        \State Extend $\Cc$ with a RWBP on $s$ referencing $\lambda_{(i,j)_s}$, $z_{(i,j)_s}$, and $\beta_{(i,j)_s}$
        \State Extend $\Qc$ for $s$ with its local connections
        \State Extend $\Ac$ for $s$ with the variable's observability if $L$
    \EndFor
    \end{algorithmic}
\end{algorithm}

\subsubsection{Update}\label{sec:rimesa:implementation-details:update}
When new factors are observed, we must not only bookkeep the new information but also update our local solution. The process of doing so is simple due to our use of the robust incremental optimizer riSAM~\cite{mcgann_risam_2023}. Encapsulating riSAM, riMESA's update is summarized in Alg.~\ref{alg:rimesa-update}.

\begin{algorithm}
\scriptsize 
\captionsetup{font=scriptsize} 
\caption{riMESA: Update (Local to robot $i$)}\label{alg:rimesa-update}
    \begin{algorithmic}[1]
    \State \textbf{In:} New factors $\Fc$ and new initial estimates $\Phi$
    \State $E_{new} \gets$ any new environment variable observed in $\Fc$
    \State $S_{new} \gets$ any new shared variable observed in $\Fc$
    \State Bookkeep($E_{new}$, $S_{new}, \Phi$, \texttt{true}) \Comment{Alg.~\ref{alg:rimesa-bookkeeping}}
    \State $\Theta \gets$  \texttt{risam.update}$(\Fc \cup \Cc, \Phi,$ \texttt{reelim=}$\Kc,$ \texttt{cvx=}$\Wc)$ \Comment{\cite[Alg.~4]{mcgann_risam_2023}}
    \State Reset $\Cc \gets \varnothing$ and $\Kc\gets\varnothing$ and $\Wc\gets\varnothing$
    \end{algorithmic}
\end{algorithm}

While simple, there are a number of nuances in this procedure. Firstly, by default, riSAM will only re-linearize factors for variables that have changed by more than a set threshold. However, since edge and dual variables are not considered variables by riSAM (as they are not optimized by riSAM), we must specifically indicate that factors affecting shared variables whose edge and dual variables have been modified since the last update (cached in $\Kc$) must be re-eliminated.

Secondly, riSAM will only treat factors with a robust kernel as potential outliers. We assume that users will mark such factors appropriately but that RWBPs are constructed automatically with robust kernels (Sec.~\ref{sec:rimesa:additional-details:robust-biased-priors}). Notably, riSAM will only re-convexify such factors when their variables are involved in the update as defined by the topological structure of the underlying factor-graph. In the distributed case, however, a robot does not know the topological structure of the other robots' graphs. Therefore, we take a conservative approach. When a shared variable is involved in an update (defined as when it is initialized), we mark for re-convexification all variables shared with that robot as well as the local variables connected with these shared variables. This set of variables for re-convexification is cached in $\Wc$.

Finally, we note that the decision to cache $\Cc, \Kc, \Wc$ and force their update during Alg.~\ref{alg:rimesa-update} was made to avoid repeated work that would be required by updating riSAM with this information explicitly after communication. This delays the information from appearing in the local solution, but the delay is minor given we expect updates from new measurements to be frequent. If effects from communication are required to be seen immediately, one can call the update step immediately after any communication at a computational cost.

\begin{remark}[riSAM Implementation Details]\label{remark:rimesa:risam-impl-details}
    We use an implementation of riSAM~\cite{mcgann_risam_2023} provided by the original authors.\footnote{\url{https://github.com/rpl-cmu/risam-v2}} We use riSAM's DogLeg Line Search to compute updates~\cite[Alg.~3]{mcgann_risam_2023}. For our M-Estimator, we use the SIG Kernel~\cite[Eq.~2]{mcgann_risam_2023} with a control parameter update sequence $\mu= [0.0, 0.5, 0.9, 0.95, 1.0]$ and a shape parameter computed such that for $\mu=1$, factors with residual $r \geq \chi^2(0.95)$ will have an influence $\left(\partial/\partial r^2\right) \rho_\mu(r^2) \leq 0.1$ to ensure that factors classified as outliers are unable to negatively affect the solution.
\end{remark}

\subsubsection{Communication}\label{sec:rimesa:implementation-details:communication}
When communication is available between two robots, C-ADMM only requires that the pair share their current solution for all shared variables. This requires minimal bandwidth on the communication link between the pair. However, the set of shared variables may have changed since the pair's last communication, and each may require initialization for some shared variables, requiring additional (though small) communication overhead. Importantly,  we must perform all of this communication in a way that is resilient to latency and communication failures (Sec.~\ref{sec:rimesa:communication-overview}).

Therefore, riMESA makes use of a two-stage communication process orchestrated by a communication handler in a separate thread. When communication is initialized between two robots, each constructs a cached version of the algorithm's internal state $\Hfr$. We denote any component of a cache with a hat $(\hat{~})$ to differentiate it from the algorithm state. First, robots update each-other on their set of shared variables, which variables they need to initialize, and their local observability for any such variable. Robots then share their local estimates for the, now jointly known, set of shared variables. This two-stage communication between robots is summarized in Alg.~\ref{alg:rimesa-communication-handler}.

\begin{algorithm}
\scriptsize 
\captionsetup{font=scriptsize} 
\caption{riMESA: Communication Handler (Robots $i \leftrightarrow j$)}\label{alg:rimesa-communication-handler}
    \begin{algorithmic}[1]
    \State \textbf{In:} State cache $\Hfr = \left\{\hat{\Theta}, \hat{S_j}, \hat{E}, \hat{\Yc_j}, \hat{\Ac}_j \triangleq \Ac\left[\hat{\Yc_j}\right] \right\}$
    \State \textbf{Out:} Communication result $\Rfr$
    \State Send $\hat{S_j}$, $\hat{E}$, $\hat{\Yc_j}$, $\hat{\Ac_j}$ to $j$ and receive $\hat{S_i}$, $\hat{E}'$, $\hat{\Yc_i}$, $\hat{\Ac_i}$  from j \Comment{Stage 1}
    \State Send $\hat{\theta}_{s_i}$ and receive $\hat{\theta}_{s_j}~\forall~s\in \hat{S_j} \cup \hat{S_i} \cup (\hat{E} \cap \hat{E}') $ \Comment{Stage 2}
    \State $\Rfr \gets$  all data sent and received
    \end{algorithmic}
\end{algorithm}

\begin{remark}[Two-Stage Communication]\label{remark:rimesa-two-stage-comms}
    A two stage communication process is required for efficient communication by any incremental C-SLAM back-end that shares state data (e.g. DDF-SAM2, DLGBP). In incremental scenarios, shared state may always change between communications. Therefore, before communicating state data, robots must first agree upon a joint set of shared variables. Alternatively, robots could simply send their entire solution, but doing so would induce a much greater communication cost.
\end{remark}

When a communication is successfully completed, a communication result $\Rfr$ containing all the data transferred in the communication is passed back to the main thread, and the information is incorporated by Alg.~\ref{alg:rimesa-incorporate-communication} to initialize new variables and update $S_j$, $\Lambda$, $Z$, and $B$. While this process is still a blocking process with respect to measurement updates (Alg.~\ref{alg:rimesa-update}), it is very efficient, only requiring writing of data and computing closed-form solutions to edge variable updates.

\begin{algorithm}
\scriptsize 
\captionsetup{font=scriptsize} 
\caption{riMESA: Incorporate Communication (Local to robot $i$)}\label{alg:rimesa-incorporate-communication}
    \begin{algorithmic}[1]
    \State \textbf{In:} Comm. Result $\Rfr = \{ \{\hat{\theta}_{s_i}\}, \hat{S_j}, \hat{E}, \hat{\Yc_j}, \hat{\Ac}_j$ and $\{\hat{\theta}_{s_j}\}, \hat{S_i}, \hat{E}', \hat{\Yc_i}, \hat{\Ac}_i\}$
    \State $\Sfr \gets \hat{S_j} \cup \hat{S_i} \cup (\hat{E} \cap \hat{E}')$ \Comment{All jointly known shared variables}
    \State $S_{new} \gets \Sfr \setminus \hat{S_j}$
    \State Bookkeep($\varnothing$, $S_{new}$, $\Theta$, \texttt{false}) \Comment{Alg.~\ref{alg:rimesa-bookkeeping}}
    \State Update $\theta_{s_j}$ and $\hat{\theta}_{s_i}$ with Init$\left(\hat{\theta}_{s_i}, \hat{\theta}_{s_j}, \hat{\Ac_j}[s]\right) ~\forall~s \in \hat{\Yc_j}$ \Comment{$i$ Robust Init.}
    \State Update $\hat{\theta}_{s_j}$ with Init$\left(\hat{\theta}_{s_i}, \hat{\theta}_{s_j}, \hat{\Ac_i}[s]\right) ~\forall~s \in \hat{\Yc_i}$ \Comment{$j$ Robust Init.}
    \State $\Yc_j \gets \Yc_j \setminus \hat{\Yc_i}$
    \For{$s$ in $\Sfr $}
        \State Update $z_{(i,j)_s} \in Z$ using $\hat{\theta}_{s_i}$, $\hat{\theta}_{s_j}$ and Eq.~\eqref{eq:mesa+-updates-z}
        \State Update $\lambda_{(i,j)_s} \in \Lambda$ using $\hat{\theta}_{s_i}$, $z_{(i,j)_s}$ and Eq.~\eqref{eq:rimesa-updates-dual-with-decay}
        \State \textbf{If} $\beta_{(i,j)_s} == \beta_{uninit}$ \textbf{then} update $B$ using $\beta_{(i,j)_s} \gets \beta_{init}$
    \EndFor
    \State Extend $\Kc$ with $\Sfr $
    \State Extend $\Wc$ with $\Sfr  \cup Q\left[\Sfr \right]$ if $\hat{\Yc_j} \neq \varnothing$
    \end{algorithmic}
\end{algorithm}

\begin{remark}[Use of Cached Data in Communication Incorporation]
We specifically call out the fact that in Alg.~\ref{alg:rimesa-incorporate-communication} all new estimates (i.e. $z_{(i,j)_s}$ and $\lambda_{(i,j)_s}$) are computed using cached data in the communication result $\Rfr$. As the actual communication (Alg.~\ref{alg:rimesa-communication-handler}) occurs in parallel, additional measurement updates may have changed the algorithm's state by the time a communication is completed. To ensure we compute edge variables that match for both robots involved in the communication, riMESA must use the values that were transmitted even if their local values have since been updated.
\end{remark}

\subsubsection{riMESA Summary}
The riMESA algorithm consists entirely of running Alg.~\ref{alg:rimesa-update} when new measurements are added, Alg.~\ref{alg:rimesa-communication-handler} when communications are initialized, and Alg.~\ref{alg:rimesa-incorporate-communication} when communications are successfully completed. If communications fail for any reason (e.g. timeout, loss of connection), the thread running Alg.~\ref{alg:rimesa-communication-handler} can simply be stopped and the attempt to communicate effectively ``discarded.'' Together these steps allow for riMESA to compute accurate solutions to C-SLAM problems in real-time even when measurements are affected by outliers and communication is ad-hoc, sparse, and unreliable.

\begin{remark}[Relationship to Prior Work]\label{remark:rimesa-relationship-to-prior-work}
iDFGO~\cite{idfgo_matsuka_2023} is an incremental factor-graph optimizer that can be applied to C-SLAM. iDFGO, however, assumes agents iterate multiple times per-step each requiring synchronized communication on a connected network, violating the communication requirements outlined in Sec.~\ref{sec:problem-definition:cslam-communication}. Additionally, iDFGO focuses on convex problems using chordal constraints for nonlinear problems, degrading performance on C-SLAM compared to geodesic constraints~\cite{mesa_mcgann_2024}. Finally, iDFGO utilizes Huber kernels for robustness, leaving it sensitive to outliers.

riMESA is a direct extension of our prior work, iMESA~\cite{imesa_mcgann_2024}. riMESA surpasses iMESA by handling outlier measurements, supporting non-blocking communications, addressing environment variables, and providing a mechanism for shared variable initialization.
\end{remark}

\section{Convergence Guarantees}\label{sec:convergence-guarantees}

We next discuss the theoretical convergence provided by C-ADMM-based C-SLAM optimizers like riMESA.

For batch optimization, variants of C-ADMM are proven to converge for non-convex problems~\cite{admm_nonconvex_converge_hong_2016, seminal_admm_nonconvex_converge_wang_2019}, on-manifold problems~\cite{riemannian_admm_li_2023}, problems with nonlinear constraints~\cite{nonlinear_constraints_converge_sun_2023}, asynchronous problems~\cite{o1k_convergence_wei_2013}, and separable problems~\cite{sova_shorinwa_2020}. However, to the best of our knowledge, C-ADMM has not been shown to converge for problems like C-SLAM that exhibit all of these traits and utilize coupled nonlinear constraints. These recent works on C-ADMM do provide promising indications that convergence could be proven for algorithms like MESA+ on batch C-SLAM. However, it also indicates that such convergence will be provable only under restrictive conditions to which real-world C-SLAM problems may not adhere. 

Moreover, to operate incrementally, riMESA introduces a number of relaxations, namely, amortizing communication, adding RWBPs, and decaying dual variables. These relaxations are necessary for riMESA to achieve quality performance in the incremental setting where the underlying problem is continuously changing, though they diverge from standard C-ADMM that assumes a fixed problem. Taking their limit (i.e. communications at time $t \rightarrow \infty$, RWBP $\rightarrow$ Weighted Biased Prior, and $\mathfrak{d} \rightarrow 1$), we recover a robust version of MESA+. This is aligned with standard C-ADMM but would be unable to operate incrementally, making the relaxations analogous to the wildfire threshold used to incrementalize standard NLS for single-robot SLAM in the iSAM2 algorithm~\cite{isam2_kaess_2012}. However, like in iSAM2, these relaxations likely preclude any formal convergence guarantees for the riMESA algorithm even if it were shown for the batch MESA+ algorithm.

While the lack of convergence guarantees is a notable limitation of riMESA, it is also standard among incremental distributed C-SLAM algorithms. DDF-SAM2 does not constrain robots to maintain equal linearization points for shared variables and in turn is guaranteed not to converge to a single consistent solution~\cite{ddfsam2_cunningham_2013}. Additionally, being an extension of Loopy Belief Propagation, DLGBP is not guaranteed to converge for even some convex problems, let alone non-convex C-SLAM problems~\cite{loopy_bp_murphy_1999}. Under these conditions we opt to focus on comparing the empirical performance of incremental distributed C-SLAM algorithms in the following section.

\section{Experiments}\label{sec:rimesa-experiments}

In this section we evaluate the performance of riMESA on a variety of real and synthetic C-SLAM problems. We demonstrate that riMESA is able to achieve superior performance compared to state-of-the-art robust, incremental, distributed C-SLAM back-ends and, through ablations, validate its design.

\subsection{Experiment Design}\label{sec:rimesa-experiments:experiment-design}
We begin by discussing the design of the experiments -- outlining the prior works and baselines to which we compare performance, the metrics for performance evaluation, the method used to generate synthetic datasets, and the model for simulated communications between robots.

\subsubsection{Prior Works \& Baselines}\label{sec:rimesa-experiments:experiment-design:prior_works_and_baselines}

A single prior work, DLGBP~\cite{robot_web_journal_murai_2024}, specifically targets robust incremental distributed C-SLAM. As discussed in Sec.~\ref{sec:related-work}, DLGBP is based on Loopy Belief Propagation and takes an M-Estimation approach to outliers, wrapping potential outlier measurements in robust kernels. We use an implementation of DLGBP provided by the authors that uses kernel parameters proposed in the original work and a window size $w=30$. 

To provide a holistic comparison, we additionally compare to a version of DDF-SAM2 ~\cite{ddfsam2_cunningham_2013} that utilizes M-Estimation for robustness. Though not proposed in the original work, the use of M-Estimators is a standard approach to enable robustness, and we argue that their use is a reasonable extension to the method. DDF-SAM2 was implemented by the authors of this work using the Naive-Bayes approximation proposed in the original paper, and potential outliers are wrapped with Geman-McClure kernels that use a shape parameter $c=3$.

In a similar vein, we also compare to a variant of riMESA that utilizes iSAM2 and M-Estimators rather than riSAM for local optimization. We refer to this variant as ``kiMESA'' due to its use of ``kernels.'' Comparison to this method is used to validate the use of riSAM for robust local optimization, making it, effectively, an ablative comparison. kiMESA makes use of Geman McClure kernels with a shape parameter $c=6$.\footnote{We also evaluated kiMESA using the same shape parameter as riMESA (Remark~\ref{remark:rimesa:risam-impl-details}). However, it was consistently outperformed by the fixed value, and we opt to omit its results for improved readability.}

In addition to these prior works, we also validate riMESA against a number of baselines. First is ``Centralized Oracle'' -- a centralized algorithm that uses batch Levenberg-Marquardt optimization and oracle outlier information (i.e. incorporates only ground-truth inlier measurements). By utilizing oracle information, this baseline provides an upper-bound on performance given that it is not constrained by limited communication and has perfect measurement classifications. 

Second is ``iMESA'' -- a non-robust version of riMESA that likewise uses oracle outlier information~\cite{imesa_mcgann_2024}. This provides a measure of the performance achievable when a method must tolerate limited communication but is unaffected by outliers.

We additionally compare against two centralized baselines that utilize state-of-the-art outlier rejection techniques. ``Centralized GNC,'' which utilizes Graduated Non-Convexity~\cite{gnc_yang_2020}, and ``Centralized PCM,'' which utilizes Pairwise Consistency Maximization~\cite{pcm_mangelson_2018}. These baselines provide a measure of the performance achievable when a method must handle outliers but is not constrained by communication limits. For these methods we use GNC provided by GTSAM,\footnote{
\url{https://github.com/borglab/gtsam}
} which uses default parameters, and PCM provided by Kimera,\footnote{
\url{https://github.com/MIT-SPARK/Kimera-RPGO}
} which uses a consistency threshold equal to $\chi^2(0.95)$. Due to their computational expense, these baselines are only used for the real-world data experiments, as running them for all synthetic dataset experiments was infeasible.

Finally, we compare against an ``Independent'' baseline in which all robots use iSAM2~\cite{isam_kaess_2007} to solve their local factor-graph without inter-robot collaboration. Local outlier measurements are handled by adding a Geman-McClure robust kernel with a fixed shape parameter ($c=3$). This baseline provides a lower-bound on the performance expected by a collaborative method, given it ignores all inter-robot information.

\subsubsection{Synthetic Datasets}\label{sec:rimesa-experiments:experiment-design:dataset_gen}
Synthetic datasets are used to explore algorithm performance across problem conditions. 

These datasets are generated for multi-robot teams containing $R=6$ robots operating in 3D obstacle-free environments. Robot motion is generated by randomly sampling odometry from a categorical distribution with options of $1m$ forward motion as well as $\pm 90^\circ$ rotation around each available axis. We generally focus on ``planar`` scenarios in which robots are constrained to traverse a planar environment and remain upright. This mobility structure is the most common for robotic applications in which robots traverse a planar (e.g. warehouse) or near-planar (e.g. outdoor) environment. However, this model can also simulate other mobility scenarios.

To construct these datasets, we simulate various types of loop-closure measurements, including intra-robot loop-closures, direct inter-robot loop-closures (i.e. robot-to-robot observations), indirect inter-robot loop-closures (i.e. via a distributed loop-closure system), and landmark observations (i.e. observations of environment features). Each type of measurement is added probabilistically when a robot is within a limited observation range, and outliers are injected such that approximately 10\% to 25\% of loop-closure measurements are outliers. When generating datasets, each type of measurement can be enabled or disabled, allowing us to simulate the types of measurements generated by different C-SLAM system designs (i.e. sensor selection and front-end algorithm design). 

All measurements are constructed with a fixed Gaussian noise model, which defines both how noise is added to simulated measurement and how algorithms model noise in their optimization. The noise models for all measurements investigated contain a rotational component, a translational component, or both. We define the noise models using a standard deviation $\sigma_r$ for any rotational components and $\sigma_t$ for any translational components. For planar-constrained scenarios, we separately specify a value for the rotational yaw axis $\sigma_{rz}$ since, in many applications, this uncertainty is larger as observations of gravity or the ground constrain pitch and roll. In the experiments below, we often explore various noise models, as large noise results in problems that are generally challenging to solve, not only because of the noise but also because large noise results in odometry drift and, in-turn poor initial estimates. Small noise, on the other hand, constructs problems that are easier to solve and have better initialization.

\begin{figure}[t]
    \vspace{-0.25cm}
    \centering
    \begin{subfigure}{1.1in}
      \centering
      \includegraphics[width=1.1in, height=1.1in]{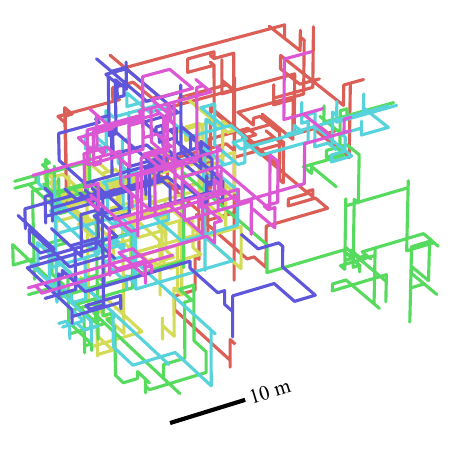}
      \vspace{-0.4cm}
      \caption{}
      \label{fig:rimesa-experiments:example-datasets:3d}
    \end{subfigure}%
    \begin{subfigure}{0.32\linewidth}
      \centering
      \includegraphics[width=1.1in, height=1.1in]{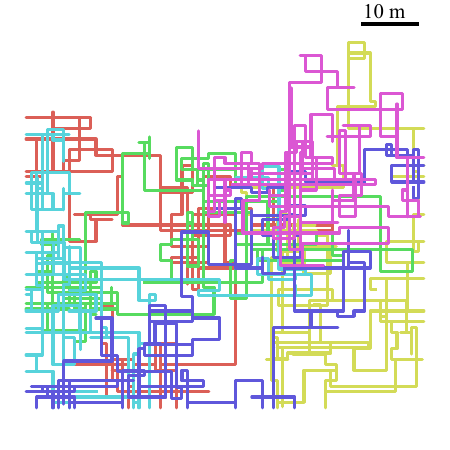}
      \vspace{-0.4cm}
      \caption{}
      \label{fig:rimesa-experiments:example-datasets:2d}
    \end{subfigure}%
    \begin{subfigure}{0.32\linewidth}
      \centering
      \includegraphics[width=1.1in, height=1.1in]{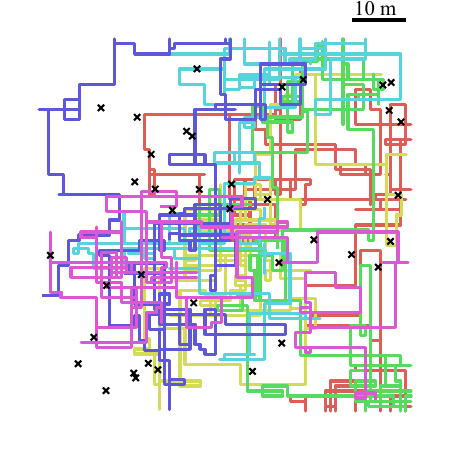}
      \vspace{-0.4cm}
      \caption{}
      \label{fig:rimesa-experiments:example-datasets:2d-landmarks}
    \end{subfigure}%
    \caption{Example groundtruth synthetic datasets. (a) An unconstrained 3D dataset. (b) A planar-constrained 3D C-PGO dataset. (c) A planar-constrained 3D landmark C-SLAM dataset. Each color represents a different robot with a trajectory length of 1000 poses.}
    \label{fig:rimesa-experiments:example-datasets}
    \vspace{-0.1cm}
\end{figure}

Examples of the synthetic datasets that can be generated by this method can be seen in Fig.~\ref{fig:rimesa-experiments:example-datasets}.

\subsubsection{Communication}\label{sec:rimesa-experiment-design:communication}
To model the communication conditions we expect in real-world deployments, we adopt the following general communication model for our experiments. We simulate communication between robots by initiating new communications at a fixed rate $r_{c}$ between any available robots within a distance of $d_{c}$ of each other. We assume that these communications are completed within a fixed period of time (resulting in a constant delay $b_{c}$) and are successful with a probability $p_{c}$. Finally, we randomly sample $5\%$ of otherwise successful communications and fail to incorporate the results for only one of the robots to simulate Two-Generals failures. 

One challenge of this model is that synthetic datasets are divorced from any understanding of time. This makes it difficult to define communication parameters, which are often measured in seconds and Hz. For experiments utilizing synthetic datasets, we adopt an abstract unit based on the dataset's odometry measurements. Specifically, we define the period between odometry measurement updates as $\mathfrak{s}$ and use this to also define a rate unit $\mathfrak{r} = 1/\mathfrak{s}$, which we use to define communication model parameters.

\begin{remark}[Synthetic Dataset Timing]\label{remark:synthetic_dataset_timing}
    Using real-world data, we can use the definitions of $\mathfrak{s}$ to derive approximate timing values to ground our synthetic datasets. An analysis of the COSMO-Bench Datasets~\cite{cosmobench_mcgann_2025} shows that they generate keyframes about every $1.5s$. Applying this to our synthetic datasets, defines $\mathfrak{r} \approx 0.67Hz$ and a 1000 pose-long trajectory as $\approx 25$ minutes of robot operation.
\end{remark}

\subsubsection{Metrics} \label{sec:rimesa-experiments:experiment-design:metrics}
We evaluate all algorithms on their capacity to produce state estimates that match the true state of the world, as this is most impactful on downstream tasks like planning and navigation. Therefore, we evaluate the performance of all methods using Average Trajectory Error (ATE) to inspect the practical quality of results~\cite{zhang_traj_metric_tutorial_2018}. We additionally evaluate the ability of an algorithm to properly classify measurements as inliers or outliers. To do so, we use the F1 score (a summary of precision and recall) computed treating ``inlier'' as the positive class. We compute ATE jointly across all robots after Umeyama alignment to the reference solution and compute the F1 score across the classifications of all robots in the team.

However, we look to evaluate these metrics not only for the final timestep, but for all intermediate timesteps during which a real multi-robot team will need accurate state estimates. To measure this performance, we use an incremental variant of these metrics~\cite{mcgann_risam_2023}. For a metric ``METRIC'' this is defined as:
\begin{equation}
\label{eq:incremental_metrics}
    \text{i\{METRIC\}} = \sum_{k \in K}\left[ \frac{k}{\sum_{k\in K} k} \text{\{METRIC\}}\left(\Theta^k, \Theta^{*}\right)\right]
\end{equation}
where $\Theta^k$ is the solution produced at iteration $k$ and $\Theta^*$ is the reference solution. As a weighted average over all timesteps, incremental metrics reward methods that maintain consistently low errors.  Compared to point metrics (e.g. metric value at the final timestep), incremental metrics provide a better measure of the performance that can be expected by downstream algorithms throughout robot operation.

For experiments with multiple trials, the incremental metrics are summarized into box-and whisker plots. For experiments with a single trial, the incremental metric is reported directly. For better readability, we report only the translation component of iATE, as the rotation and translation components are highly correlated. Finally, there are cases in the experiments where algorithms fail.\footnote{Failures include when an algorithm diverges to an unreasonable solution or when the algorithm crashes from numerical instability issues.} If trials fail, the summary box-and-whisker plot is rendered with dashed lines, and the percentage of successful trials summarized in the plot is written above.

\subsection{C-SLAM Generalization Experiment}\label{sec:rimesa-experiments:generalization}
In our first experiment, we evaluate the ability of riMESA to generalize to different problem scenarios. We design each scenario by enabling and disabling different types of measurements (Sec.~\ref{sec:rimesa-experiments:experiment-design:dataset_gen}). This is used to simulate the measurements produced by various C-SLAM system designs.

For each of these scenarios, we additionally evaluate across different levels of measurement noise to simulate different difficulties of problems. For each noise level and scenario combination, we generate 50 random datasets that simulate a team of 6 robots traversing trajectories 1000 poses long. 

For all scenarios, robots can initiate new communications at a rate of $r_c = 1\mathfrak{r}$ with any robot within a communication range of $d_c = 30m$. We assume that communications are successful with $p_c = 0.9$ and are completed before the next measurement update is received (effective delay $b_c < 1\mathfrak{s}$).

\begin{figure*}[t]
    \centering
    \begin{subfigure}{0.5\linewidth}
        \centering
        \includegraphics[width=3.45in,height=1.75in]{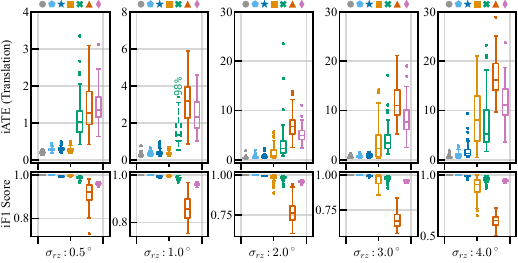}
        \vspace{-2pt}
        \caption{Collaborate Pose-Graph Optimization (C-PGO)}
        \vspace{4pt}
        \label{fig:rimesa-experiments:generalization:cpgo-planar}
    \end{subfigure}%
    \begin{subfigure}{0.5\linewidth}
        \centering
        \includegraphics[width=3.45in,height=1.75in]{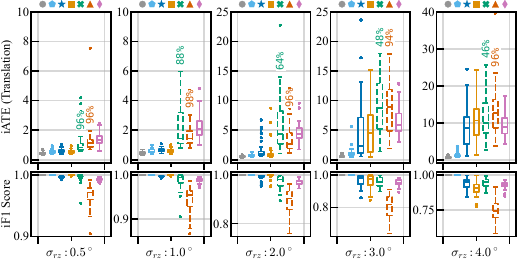}
        \vspace{-4pt}
        \caption{Range-Aided PGO}
        \vspace{4pt}
        \label{fig:rimesa-experiments:generalization:range-aided-pgo-planar}
    \end{subfigure}
    \begin{subfigure}{0.5\linewidth}
            \centering
        \includegraphics[width=3.45in,height=1.75in]{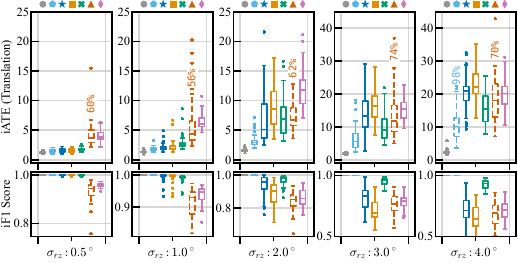}
        \vspace{-4pt}
        \caption{Range-Only C-SLAM}
        \vspace{4pt}
        \label{fig:rimesa-experiments:generalization:range-only-cslam-planar}
    \end{subfigure}%
    \begin{subfigure}{0.5\linewidth}
            \centering
        \includegraphics[width=3.45in,height=1.75in]{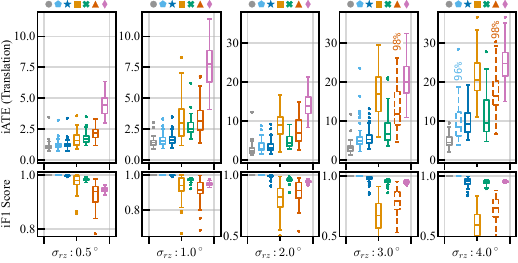}
        \vspace{-4pt}
        \caption{Bearing-Range-Only C-SLAM}
        \vspace{4pt}
        \label{fig:rimesa-experiments:generalization:bearing-range-only-cslam-planar}
    \end{subfigure}
    \begin{subfigure}{0.5\linewidth}
        \centering
        \includegraphics[width=3.45in,height=1.75in]{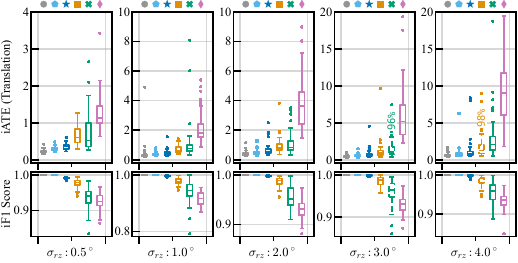}
        \vspace{-4pt}
        \caption{Landmark C-SLAM}
        \vspace{4pt}
        \label{fig:rimesa-experiments:generalization:landmark-cslam-planar}
    \end{subfigure}%
    \begin{subfigure}{0.5\linewidth}
        \centering
        \includegraphics[width=3.45in,height=1.75in]{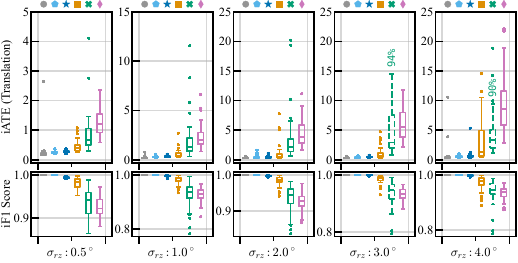}
        \vspace{-4pt}
        \caption{Landmark+Direct C-SLAM}
        \vspace{2pt}
        \label{fig:rimesa-experiments:generalization:landmark-direct-cslam-planar}
    \end{subfigure}
    \vspace{-0.7cm}
    \caption{Metric performance for Centralized Oracle~(\SymCentralized), iMESA~(\SymIMESA), riMESA~(\SymRIMESA), kiMESA~(\SymKIMESA), DDF-SAM2~(\SymDDFSAM), DLGBP~(\SymDLGBP), and Independent~(\SymIndep) on various planar-constrained datasets for different levels of measurement noise. Only the yaw axis noise ($\sigma_{rz}$) is changed, and datasets use fixed $\sigma_r=0.25^\circ$ and $\sigma_t=0.05m$. Across all problem scenarios, our proposed method, riMESA, outperforms prior works and achieves the closest solution to that of the baselines that exploit oracle outlier information. However, there are some problem conditions for which riMESA struggles at high levels of measurement noise. Note: DLGBP is omitted from landmark scenarios as it does not support passive landmarks.}
    \label{fig:rimesa-experiments:generalization}
    \vspace{-0.6cm}
\end{figure*}

\subsubsection{Collaborate PGO (C-PGO) (Fig.\ref{fig:rimesa-experiments:generalization:cpgo-planar})} We first look at a PGO scenario in which robots generate relative-pose intra-robot loop-closures as well as relative-pose indirect inter-robot loop-closures. C-PGO systems are currently particularly popular as they are mathematically well constrained and the geometric sensors used to derive relative-pose measurements (i.e. LiDARs) are readily available and highly accurate.

\subsubsection{Range-Aided PGO (Fig.\ref{fig:rimesa-experiments:generalization:range-aided-pgo-planar})} We next look at a scenario in which agents locally perform PGO and derive only direct inter-robot ranging measurements (e.g. from Ultra-Wide-Band (UWB) radios) to support collaboration.

\subsubsection{Range-Only C-SLAM (Fig.\ref{fig:rimesa-experiments:generalization:range-only-cslam-planar})} Next we look at a scenario like that above but where, due to sensor or computational limitations, robots utilize only their local odometry and direct inter-robot ranging measurements to localize the team.

\subsubsection{Bearing-Range-Only C-SLAM (Fig.\ref{fig:rimesa-experiments:generalization:bearing-range-only-cslam-planar})} Next we look at a scenario like that above but in which robots can also observe the bearing to the target teammate when ranging.

\subsubsection{Landmark C-SLAM (Fig.\ref{fig:rimesa-experiments:generalization:landmark-cslam-planar})} Next we look at a scenario in which robots utilize only odometry and measurements to landmarks in their environment. Such system design is commonly used when visual sensors are the primary sensor onboard each robot. Note that for this scenario, the prior work DLGBP is omitted as its distributed formulation is unable to support passive landmarks and instead requires landmarks to support active computation and communication.

\subsubsection{Landmark+Direct C-SLAM (Fig.\ref{fig:rimesa-experiments:generalization:landmark-direct-cslam-planar})} Next, we look at a landmark C-SLAM scenario in which robots additionally make direct inter-robot bearing-range measurements to teammates. Again, DLGBP is omitted as it requires active landmarks.

\begin{figure}[ht]
    \centering
    \includegraphics[width=3.45in,height=1.75in]{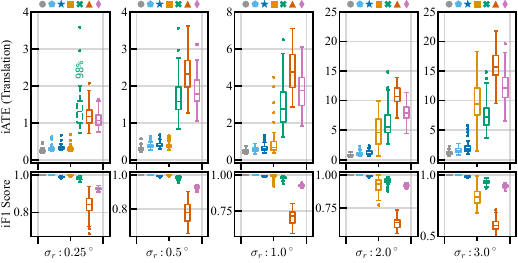}
    \vspace{-0.5cm}
    \caption{Metric performance for Centralized Oracle~(\SymCentralized), iMESA~(\SymIMESA), riMESA~(\SymRIMESA), kiMESA~(\SymKIMESA), DDF-SAM2~(\SymDDFSAM), DLGBP~(\SymDLGBP), and Independent~(\SymIndep) on a unconstrained 3D C-PGO datasets for different levels of measurement noise. Only $\sigma_{r}$ is changed and datasets use fixed $\sigma_t=0.05m$.}
    \label{fig:rimesa-experiments:generalization:cpgo-3d}
    \vspace{-0.1cm}
\end{figure}

\subsubsection{Unconstrained 3D C-PGO (Fig.~\ref{fig:rimesa-experiments:generalization:cpgo-3d}.)} Finally, we look at a scenario in which robots are not constrained to a planar environment but rather can move arbitrarily in 3D. Such scenarios are extremely rare in robotics (i.e. zero-gravity) but are included for theoretical completeness. 

\subsubsection{Generalization Experiment Analysis} \label{sec:rimesa-experiments:generalization:analysis}
In this experiment we can see that riMESA significantly outperforms prior works, producing solutions that are closest to the baselines that utilize oracle outlier information. This provides us confidence that riMESA can generalize across C-SLAM problem scenarios to adapt to the sensors and front-end algorithms that are necessary for the application. We can additionally see some scenarios where riMESA's performance degrades. Notably, riMESA struggles on the Range-Aided PGO (\ref{fig:rimesa-experiments:generalization:range-aided-pgo-planar}), Range-Only C-SLAM (\ref{fig:rimesa-experiments:generalization:range-only-cslam-planar}), and Bearing-Range-Only C-SLAM (\ref{fig:rimesa-experiments:generalization:bearing-range-only-cslam-planar}) scenarios in cases of large measurement noise. The commonality among these scenarios is that they are scenarios where the inter-robot measurements are low-rank. Therefore, in cases of large noise, these are scenarios where the effective signal-to-noise ratio is low. Interestingly, in these cases the iMESA baseline also struggles while the Centralized Oracle performs well. This indicates that it is the limited communication rather than the noisy measurements or outliers that is causing the poor performance. We, therefore, hypothesize that with more communication, riMESA can succeed in these conditions. This hypothesis is explored below in Sec.~\ref{sec:rimesa-experiments:communication}.

In this experiment we can also see that DDF-SAM2 and DLGBP struggle across most problem conditions. Both prior works can sometimes produce quality results but are just as likely to compute poor estimates or diverge entirely. kiMESA, our ablative method, often outperforms these prior works. However, as expected from findings in robust optimization literature, the use of a Geman McClure robust kernel makes it sensitive to initialization, and kiMESA struggles to converge at large noise levels even in high signal-to-noise scenarios. It is also worth noting that in these experiments there are some instances of iMESA diverging. This is likely due to its use of Gauss-Newton optimization steps. This differs from riMESA, which uses a trust-region optimizer. The use of trust-region optimization steps in iMESA would likely resolve this issue.

\subsection{Scale Experiment}\label{sec:rimesa-experiments:scale}
\begin{figure}[t]
    \centering
    \begin{subfigure}{1.0\linewidth}
        \centering
        \includegraphics[width=3.45in,height=1.75in]{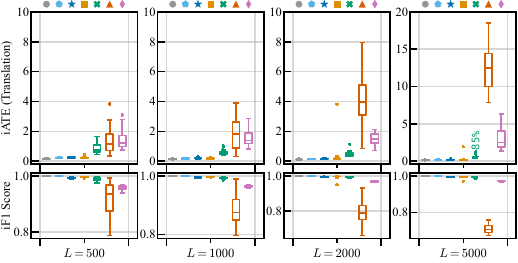}
        \caption{Long Term Operation}
        \vspace{4pt}
        \label{fig:rimesa-experiments:scale-experiment:duration}
    \end{subfigure}
    \begin{subfigure}{1.0\linewidth}
        \centering
        \includegraphics[width=3.45in,height=1.75in]{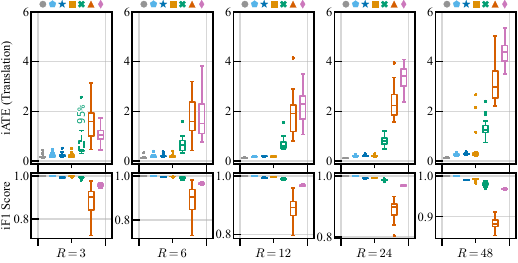}
        \caption{Team-Size}
        \label{fig:rimesa-experiments:scale-experiment:team-size}
    \end{subfigure}
    \caption{Metric performance for Centralized Oracle~(\SymCentralized), iMESA~(\SymIMESA), riMESA~(\SymRIMESA), kiMESA~(\SymKIMESA), DDF-SAM2~(\SymDDFSAM), DLGBP~(\SymDLGBP), and Independent~(\SymIndep) across different (a) operation lengths $L$ and (b) different team-sizes $R$. riMESA provides quality performance across all problem scales.}
    \label{fig:rimesa-experiments:scale-experiment}
    \vspace{-0.1cm}
\end{figure}

In our next experiment, we evaluate how riMESA scales with long-term operation and large teams of robots. In both cases we evaluate on 20 planar-constrained 3D C-PGO datasets. For the long-term operation experiment, we simulate a team with 6 robots over various trajectory lengths (L). For the team-size experiment, we simulate different robot team sizes (R), where each robot traverses a trajectory of 1000 poses. For both, we use the same communication model as in the Generalization Experiment (Sec.\ref{sec:rimesa-experiments:generalization}) and a noise model with $\sigma_{rz}=1.0^\circ$, $\sigma_r=0.25^\circ$, and $\sigma_t=0.05m$. Results from the long-term operation experiment can be found in Fig.~\ref{fig:rimesa-experiments:scale-experiment:duration}, and results from the team-size experiment can be seen in Fig.~\ref{fig:rimesa-experiments:scale-experiment:team-size}.

In this experiment we can see that riMESA is able to scale to both large teams and long-term operation, providing consistent, high-quality results across scales. Both prior works, DDF-SAM2 and DLGBP, see degraded performance at large scales. However, this is most pronounced for DLGBP, which performs particularly poorly during long-term operation. This is likely caused as DLGBP struggles to incorporate long-term loop-closure information due to windowing, which is necessary to maintain online efficiency.

\newcommand{\commquality}[3]{$[#1\mathfrak{s}, #2\mathfrak{r}, #3m]$}
\begin{figure}[t]
    \centering
    \begin{subfigure}{1\linewidth}
        \centering
        \includegraphics[width=3.45in,height=1.75in]{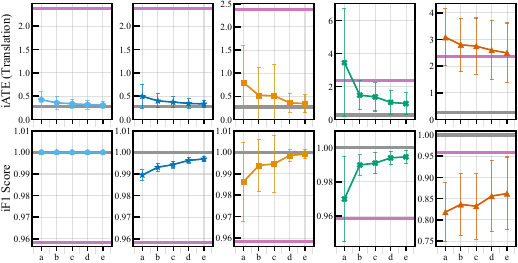}
        \caption{Collaborate Pose-Graph Optimization ($\sigma_r=0.25^\circ$, $\sigma_{rz}=1^\circ$, $\sigma_t=0.05m$)}
        \vspace{4pt}
        \label{fig:rimesa-experiments:comm-experiment:planar-pgo}
    \end{subfigure}
    \begin{subfigure}{1.0\linewidth}
        \centering
        \includegraphics[width=3.45in,height=1.75in]{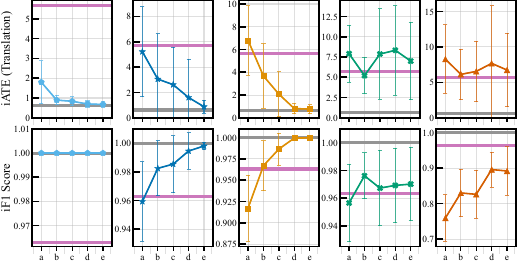}
        \caption{Range-Aided PGO ($\sigma_r=0.25^\circ$, $\sigma_{rz}=2.5^\circ$, $\sigma_t=0.05m$)}
        \vspace{4pt}
        \label{fig:rimesa-experiments:comm-experiment:planar-range-aided-pgo}
    \end{subfigure}
    \begin{subfigure}{1.0\linewidth}
        \centering
        \includegraphics[width=3.45in,height=1.75in]{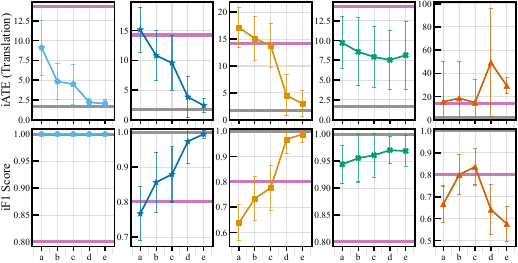}
        \caption{Range-Only C-SLAM ($\sigma_r=0.25^\circ$, $\sigma_{rz}=2.5^\circ$, $\sigma_t=0.05m$)}
        \label{fig:rimesa-experiments:comm-experiment:planar-range-only-cslam}
    \end{subfigure}%
    \caption{Metric performance for iMESA~(\SymIMESA), riMESA~(\SymRIMESA), kiMESA~(\SymKIMESA), DDF-SAM2~(\SymDDFSAM), and DLGBP~(\SymDLGBP) on various planar-constrained problem scenarios for different communication conditions. The baselines Centralized Oracle~(\SymCentralized) and Independent~(\SymIndep) are not dependent on communication and are plotted as horizontal lines. Communication quality, defined by [delay ($b_c$), rate ($r_c$), max-distance ($d_c$)], improves from a to e, where a=\commquality{10}{1}{30}, b=\commquality{2}{1}{35}, c=\commquality{0}{1}{40}, d=\commquality{0}{5}{45}, e=\commquality{0}{10}{50}. Higher quality communication enables riMESA to provide better results across all scenarios, but particularly for low signal-to-noise problems like (c) Range-Only SLAM.}
    \label{fig:rimesa-experiments:comm-experiment}
    \vspace{-0.1cm}
\end{figure}

\subsection{Communication Experiments} \label{sec:rimesa-experiments:communication}
In our next experiment, we seek to evaluate how the quality of communication affects riMESA. We define ``poor'' communication quality as when robots have a limited communication range $r_c$ and can only communicate with long delays $b_c$ due to latency and limited bandwidth. Likewise, we define ``good'' communication quality as when robots can communicate over long ranges at high effective rates.

For ``poor'' communication scenarios, we explicitly simulate delays to test algorithms' abilities to tolerate receiving stale data. In these scenarios robots still attempt to initiate new communications at a faster rate, and we permit robots to maintain communications with multiple robots at once.\footnote{The implementation of DLGBP provided by the original authors does not permit us to simulate delays. Thus, while we affect the rate of communication for DLGBP, we are unable to simulate incorporation of stale data.}

We look at the effect of communication under three different scenarios: a standard C-PGO scenario, a Range-Aided PGO scenario, and a Range-Only C-SLAM scenario. C-PGO was selected as a representative generic scenario. The others were selected due to our observations in the Generalization Experiment (Sec.~\ref{sec:rimesa-experiments:generalization}) that -- at larger noise levels, riMESA's performance on these scenarios significantly degrades. We explore these scenarios to investigate if higher quality communication can enable quality performance in these challenging scenarios. For these scenarios, we specifically select noise levels at which riMESA's performance begins to degrade. For each combination of communication quality and problem scenario, we evaluate 20 random synthetic datasets. Results from this experiment can be seen in Fig.~\ref{fig:rimesa-experiments:comm-experiment}.

In this experiment we can see that, for high signal-to-noise problems (Fig.~\ref{fig:rimesa-experiments:comm-experiment:planar-pgo}), riMESA is robust to communication conditions, producing quality results even with very limited communication. We can further see that for low signal-to-noise problem scenarios (Fig.~\ref{fig:rimesa-experiments:comm-experiment:planar-range-aided-pgo} and Fig.~\ref{fig:rimesa-experiments:comm-experiment:planar-range-only-cslam}), improvements to communication quality significantly improve results. This supports the hypothesis introduced in Sec.~\ref{sec:rimesa-experiments:generalization:analysis} -- that riMESA can perform well under low signal-to-noise conditions but requires more communication to achieve quality results.

We can also see that iMESA performs better with higher quality communication, though, compared to riMESA, the improvements are far less drastic. This indicates that while our C-ADMM-based design allows outliers to be handled locally from a mathematical perspective, practically the handling of outliers still imposes some communication burden onto the back-end. Retrospectively, this makes intuitive sense. The robust problem is significantly more complex, and it is reasonable that a multi-robot team will require more communication to converge to a quality solution on a more complex problem.

Additionally, this experiment shows that while iMESA, riMESA, and kiMESA all improve their performance with higher quality communication, the prior works DLGBP and DDF-SAM2 do not see the same improvements. For some problem scenarios, more communication does little to affect the results of these prior works at all. Finally, it is worth noting that while higher quality communication allows kiMESA to improve performance, we still expect this method to struggle in scenarios with large noise, as seen in Fig.~\ref{fig:rimesa-experiments:generalization}.

\subsection{Ablation Experiments}\label{sec:rimesa-experiments:ablations}
We next perform some ablative tests to validate our design decisions for riMESA's robust optimization technique (Sec.~\ref{sec:rimesa:robust-distribued-opt-overview}) and robust initialization scheme (Sec.~\ref{sec:rimesa:additional-details:shared-variable-initialization}).

\subsubsection{Robust Optimization Ablation} \label{sec:rimesa-experiments:ablations:robust-opt}
\begin{figure}[t]
    \begin{subfigure}{1in}
        \centering
        \includegraphics[width=1in, height=1.75in]{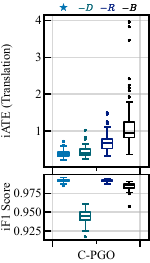}
        \caption{}
        \label{fig:rimesa-experiments:abaltion:robust-opt}
    \end{subfigure}%
    \hspace{0.2in}
    \begin{subfigure}{2.2in}
        \centering
        \includegraphics[width=2.20in, height=1.75in]{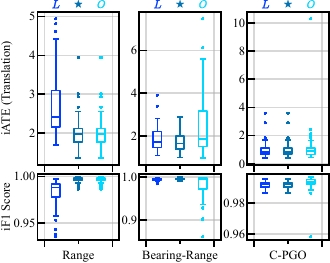}
        \caption{}
        \label{fig:rimesa-experiments:abaltion:robust-init}
    \end{subfigure}
    \caption{riMESA ablation studies. In (a) we compare riMESA~(\SymRIMESA) to a version without dual-decay (\textcolor{sns:dark:light_blue}{\textit{\textbf{--D}}}), a version without robust kernels on weighted biased priors (\textcolor{sns:dark:blue}{\textit{\textbf{--R}}}), and a version without both of these components (\textit{\textbf{--B}}). In (b) we compare riMESA~(\SymRIMESA) to a version that uses only local estimates to initialize shared variables (\textcolor{sns:bright:blue}{\textit{\textbf{L}}}) and a version that uses only estimates from owner robots to initialize shared variables (\textcolor{sns:bright:light_blue}{\textit{\textbf{O}}}). Both ablation studies validate that the components improve the performance of riMESA.}
    \label{fig:rimesa-experiments:abaltion}
    \vspace{-0.1cm}
\end{figure}

In this experiment we look at the removal of two components of our robust optimization scheme. Namely, we individually remove the wrapping of weighted biased priors in robust kernels (``--R''), remove use of a decay rate for dual variables (``--D''), and remove both (`--B''). We compare these variants against riMESA on 50 planar 3D C-PGO problems and use the same communication model as the Generalization Experiment (Sec.\ref{sec:rimesa-experiments:generalization}) and a noise model with $\sigma_{rz}=1.0^\circ$, $\sigma_r=0.25^\circ$, and $\sigma_t=0.05m$. The results from this experiment can be seen in Fig.~\ref{fig:rimesa-experiments:abaltion:robust-opt} and show that the use of dual-decay and robust weighted biased priors significantly improves the performance of riMESA.

\subsubsection{Robust Initialization Ablation}\label{sec:rimesa-experiments:ablations:robust-init}
In this experiment we analyze the effect of our robust initialization scheme. We compare the robust initialization scheme against a variant that uses only local information and default values to initialize shared variables (``L'') and a variant that uses the estimate of the owner robot to initialize shared variables (``O''). We compare these variants on three scenarios with different local observability -- a Range-Only C-SLAM scenario in which robots do not directly observe any shared variable state locally, a Bearing-Range-Only C-SLAM scenario in which robots locally observe shared variable locations, and a C-PGO scenario in which robots locally observe the entire shared variable state. 

For each scenario, we evaluate 50 planar-constrained problems and use the same communication model as the Generalization Experiment (Sec.\ref{sec:rimesa-experiments:generalization}). For all scenarios, we use noise model parameters $\sigma_r=0.25^\circ$ and $\sigma_t=0.05m$. For the C-PGO scenario, we define $\sigma_{rz}=3.0^\circ$, and for the range and bearing-range scenarios, we define $\sigma_{rz}=1.0^\circ$. 

The results from this ablation study can be seen in Fig.~\ref{fig:rimesa-experiments:abaltion:robust-init} and show that the robust initialization scheme always results in the best overall performance. It is worth explicitly noting that the robust initialization scheme is equivalent to other approaches in some scenarios. For example, in the range-only scenarios, the robust initialization approach is equivalent to using the owner robot's estimate for initialization.

\subsection{Real-World Experiments}\label{sec:rimesa-experiments:cosmobench}
In our final experiment, we seek to evaluate the performance of riMESA on data that (as closely as possible) resembles real-world data. To achieve this goal while supporting reproduction of our results, we adopt COSMO-Bench~\cite{cosmobench_mcgann_2025}. This benchmark consists of 24 datasets generated from real-world LiDAR data~\cite{mcd_nguyen_2024, cumulti_albin_2025}, a baseline LiDAR-based C-SLAM front-end, and a communication model derived from real-world data. It also provides four additional datasets from the Nebula dataset~\cite{lamp2_chang_2022}.

For this experiment, we adopt communication models based on those proposed by COSMO-Bench. Specifically, we assume that robots initiate new communications at a rate of $r_c = 5Hz$, that the time required to complete each communication is $b_c \leq 200ms$, and that the network bandwidth is always sufficient to transmit the data required by the algorithms. We model communication success $p_c$ and max communication range $d_c$ using the connectivity model (i.e. Wi-Fi or Pro-Radio) applicable to the dataset~\cite[Sec.IV.C]{cosmobench_mcgann_2025}.\footnote{For the Nebula datasets, we use the Pro-Radio connectivity model.} We additionally continue to sample a small fraction ($5\%$) of otherwise successful communications to simulate Two-Generals failures by failing to incorporate the results for one of the robots.

\begin{table*}[t] 
    \centering
    \scriptsize
    \renewcommand*{\arraystretch}{1.2}
    \caption{iATE and Final ATE performance for all methods and baselines on the COSMO-Bench Datasets~\cite{cosmobench_mcgann_2025}. The best-performing method (excluding baselines) for each dataset is bolded. The proposed algorithm, riMESA (highlighted in gray), performs consistently across all datasets.}
    \vspace{-0.2cm}
    \label{tab:rimesa-experiments:cosmobench:main:quantitative}
\newcommand{\header}[1]{\rotatebox[origin=c]{90}{\texttt{\parbox{1.4cm}{\centering #1}}}}
\newcolumntype{Y}{>{\centering\arraybackslash}X}
\begin{tabularx}{\textwidth}{|c|c|l|Y|Y|Y|Y|Y|Y|Y|Y|Y|Y|Y|Y||Y|Y|Y|Y|Y|Y|Y|Y|Y|Y|Y|Y|}
\hline
\multicolumn{3}{|c|}{Metric} & \multicolumn{12}{c||}{iATE (m)} & \multicolumn{12}{c|}{Final ATE (m)} \\
\hline
\multicolumn{3}{|c|}{Trial} & \header{kth\\r3\_00} & \header{kth\\r3\_01} &  \header{kth\\r4\_00} &  \header{ntu\\r3\_00} &  \header{ntu\\r3\_01} &  \header{ntu\\r3\_02} &  \header{ntu\\r4\_00} &  \header{ntu\\r5\_00} &  \header{tuhh\\r3\_00} &  \header{tuhh\\r3\_01} &  \header{kittredge loop} &  \header{main campus} & \header{kth\\r3\_00} & \header{kth\\r3\_01} &  \header{kth\\r4\_00} &  \header{ntu\\r3\_00} &  \header{ntu\\r3\_01} &  \header{ntu\\r3\_02} &  \header{ntu\\r4\_00} &  \header{ntu\\r5\_00} &  \header{tuhh\\r3\_00} &  \header{tuhh\\r3\_01} &  \header{kittredge loop} &  \header{main campus}  \\
\hline

\multirow{9}{*}[1.5ex]{\rotatebox[origin=c]{90}{Wi-Fi Datasets}}
&\multirow{6}{*}[1.5ex]{\rotatebox[origin=c]{90}{Baselines}}
& \SymIndep~Independent &8.24 & 11.5 & 8.76 & 6.99 & 7.51 & 5.93 & 7.94 & 7.50 & 17.1 & 15.4 & 33.2 & 18.2 & 9.52 & 13.1 & 8.67 & 7.05 & 8.45 & 7.98 & 8.25 & 7.73 & 19.0 & 19.4 & 47.4 & 16.9\\
&& \SymCentralized~Cent. Oracle &2.83 & 4.51 & 2.88 & 2.28 & 2.01 & 4.44 & 1.79 & 1.94 & 1.96 & 4.42 & 5.47 & 8.53 & 1.91 & 1.52 & 1.19 & 1.65 & 2.08 & 1.81 & 1.37 & 1.39 & 1.10 & 1.15 & 4.58 & 3.83\\
&& \SymIMESA~iMESA &5.51 & 5.10 & 4.72 & 2.72 & 3.34 & 5.24 & 2.58 & 2.52 & 6.31 & 3.99 & 5.93 & 9.36 & 2.54 & 2.96 & 4.12 & 1.62 & 3.20 & 2.46 & 1.63 & 1.53 & 2.80 & 1.04 & 4.90 & 3.95\\
&& \SymGNC~Cent. GNC &5.42 & 5.25 & 3.23 & 2.53 & 2.36 & 4.53 & 2.10 & 2.32 & 1.86 & 5.19 & 7.41 & 9.98 & 2.16 & 1.51 & 1.30 & 2.06 & 2.51 & 3.67 & 1.85 & 1.80 & 0.85 & 1.23 & 5.06 & 3.83\\
&& \SymPCM~Cent. PCM &28.7 & 43.3 & 80.1 & 2.56 & 2.27 & 4.49 & 2.15 & 2.19 & 3.14 & 4.74 & 25.9 & 13.1 & 79.7 & 80.1 & 77.9 & 1.84 & 2.39 & 1.81 & 1.70 & 1.52 & 1.43 & 1.46 & 87.1 & 5.09\\
\cline{2-27}
&\multirow{4}{*}[0ex]{\rotatebox[origin=c]{90}{Methods}}
& \SymDDFSAM~DDF-SAM2 &11.4 & 53.3 & 26.1 & \textbf{3.27} & 23.8 & 4.81 & \textbf{2.60} & 21.9 & 98.9 & 4.81 & \textbf{10.8} & 12.8 & 9.27 & 86.4 & 55.7 & \textbf{1.90} & 22.7 & \textbf{2.26} & \textbf{1.58} & 25.2 & 33.7 & 5.37 & \textbf{5.38} & \textbf{3.74}\\
&& \SymDLGBP~DLGBP &6.80 & 13.0 & 11.9 & 11.6 & 8.27 & 4.88 & 10.9 & 10.9 & 12.2 & 17.8 & 71.9 & 58.1 & 7.46 & 15.1 & 13.0 & 13.5 & 12.0 & 6.41 & 13.3 & 13.0 & 21.2 & 28.1 & 99.3 & 76.4\\
&& \SymKIMESA~kiMESA &6.60 & 11.8 & 9.65 & 5.14 & 4.56 & 5.56 & 3.10 & 5.63 & 10.5 & 29.7 & 34.9 & 15.3 & 6.36 & 12.0 & 9.09 & 2.10 & 4.48 & 7.07 & 2.19 & 5.30 & 7.21 & 62.2 & 8.06 & 3.77\\
&&\cellcolor{gray!20}\SymRIMESA~riMESA &\cellcolor{gray!20} \textbf{4.46} &\cellcolor{gray!20}  \textbf{4.56} &\cellcolor{gray!20}  \textbf{4.30} &\cellcolor{gray!20}  3.84 &\cellcolor{gray!20}  \textbf{3.27} &\cellcolor{gray!20}  \textbf{4.50} &\cellcolor{gray!20}  4.72 &\cellcolor{gray!20}  \textbf{3.92} &\cellcolor{gray!20}  \textbf{2.66} &\cellcolor{gray!20}  \textbf{3.91} &\cellcolor{gray!20}  16.8 &\cellcolor{gray!20}  \textbf{10.3} &\cellcolor{gray!20}  \textbf{2.58} &\cellcolor{gray!20}  \textbf{2.28} &\cellcolor{gray!20}  \textbf{3.56} &\cellcolor{gray!20}  2.95 &\cellcolor{gray!20}  \textbf{3.04} &\cellcolor{gray!20}  3.22 &\cellcolor{gray!20}  2.96 &\cellcolor{gray!20}  \textbf{2.95} &\cellcolor{gray!20}  \textbf{1.59} &\cellcolor{gray!20}  \textbf{1.27} &\cellcolor{gray!20}  7.34 &\cellcolor{gray!20}  4.29\\
\hline\multicolumn{27}{c}{\vspace{-6pt}}\\\hline

\multirow{9}{*}[1.5ex]{\rotatebox[origin=c]{90}{Pro-Radio Datasets}}
&\multirow{6}{*}[1.5ex]{\rotatebox[origin=c]{90}{Baselines}}
& \SymIndep~Independent &7.33 & 12.5 & 11.2 & 7.14 & 7.22 & 5.76 & 7.69 & 7.81 & 16.5 & 10.9 & 29.9 & 35.5 & 8.19 & 14.3 & 12.0 & 7.45 & 7.82 & 7.99 & 7.42 & 8.09 & 18.7 & 15.7 & 36.8 & 23.9\\
&& \SymCentralized~Cent. Oracle &2.39 & 3.03 & 2.58 & 1.59 & 1.53 & 1.97 & 1.92 & 1.86 & 1.71 & 1.37 & 4.72 & 4.40 & 1.13 & 1.42 & 1.25 & 1.64 & 1.42 & 1.78 & 1.46 & 1.50 & 1.24 & 1.26 & 4.28 & 3.20\\
&& \SymIMESA~iMESA &6.68 & 3.81 & 4.01 & 1.63 & 1.66 & 2.02 & 2.51 & 2.20 & 5.36 & 2.17 & 5.01 & 5.03 & 3.61 & 3.07 & 1.90 & 1.47 & 1.50 & 1.89 & 1.75 & 1.68 & 3.49 & 1.07 & 4.56 & 3.47\\
&& \SymGNC~Cent. GNC &3.73 & 2.87 & 2.77 & 2.04 & 2.07 & 2.50 & 2.28 & 1.99 & 1.59 & 1.67 & 5.56 & 4.69 & 1.25 & 1.53 & 1.33 & 2.16 & 2.14 & 2.97 & 2.10 & 1.81 & 1.27 & 1.32 & 4.67 & 3.25\\
&& \SymPCM~Cent. PCM &72.5 & 57.4 & 72.0 & 2.05 & 14.7 & 1.98 & 2.17 & 2.12 & 79.8 & 3.86 & 10.9 & 6.74 & 60.4 & 60.2 & 47.0 & 1.77 & 5.39 & 1.75 & 1.68 & 1.63 & 105.7 & 1.38 & 6.05 & 5.26\\
\cline{2-27}
&\multirow{4}{*}[0ex]{\rotatebox[origin=c]{90}{Methods}}
& \SymDDFSAM~DDF-SAM2 &45.2 & 98.8 & 40.4 & 5.38 & 1.77 & \textbf{1.89} & 5.73 & 3.48 & 53.8 & 36.5 & 32.6 & 12.8 & 54.4 & 101.5 & 42.3 & \textbf{1.67} & \textbf{1.64} & \textbf{1.65} & \textbf{1.69} & 1.83 & 30.1 & 22.5 & 13.4 & \textbf{3.50}\\
&& \SymDLGBP~DLGBP &6.32 & 13.0 & 11.6 & 11.3 & 8.40 & 4.78 & 12.0 & 11.3 & 8.06 & 12.5 & 72.4 & 57.7 & 6.88 & 15.1 & 12.8 & 13.4 & 12.5 & 6.27 & 14.7 & 13.5 & 17.0 & 25.5 & 100.3 & 75.9\\
&& \SymKIMESA~kiMESA &10.7 & \textbf{3.55} & 12.1 & \textbf{1.89} & \textbf{1.75} & 2.59 & 5.19 & \textbf{2.29} & 7.62 & 5.29 & 32.8 & 34.3 & 11.5 & \textbf{2.58} & 12.3 & 1.81 & 1.66 & 2.76 & 4.95 & \textbf{1.78} & 3.96 & 3.72 & 31.2 & 22.9\\
&&\cellcolor{gray!20}\SymRIMESA~riMESA &\cellcolor{gray!20} \textbf{4.97} &\cellcolor{gray!20} 3.67 &\cellcolor{gray!20} \textbf{4.33} &\cellcolor{gray!20} 3.14 &\cellcolor{gray!20} 4.85 &\cellcolor{gray!20} 2.99 &\cellcolor{gray!20} \textbf{4.43} &\cellcolor{gray!20} 3.25 &\cellcolor{gray!20} \textbf{5.36} &\cellcolor{gray!20} \textbf{2.20} &\cellcolor{gray!20} \textbf{6.96} &\cellcolor{gray!20} \textbf{8.15} &\cellcolor{gray!20} \textbf{3.51} &\cellcolor{gray!20} 3.17 &\cellcolor{gray!20} \textbf{2.58} &\cellcolor{gray!20} 2.96 &\cellcolor{gray!20} 3.44 &\cellcolor{gray!20} 3.69 &\cellcolor{gray!20} 4.09 &\cellcolor{gray!20} 3.16 &\cellcolor{gray!20} \textbf{3.58} &\cellcolor{gray!20} \textbf{1.34} &\cellcolor{gray!20} \textbf{5.77} &\cellcolor{gray!20} 3.84\\
\hline
\end{tabularx}

    \vspace{-0.5cm}
\end{table*}

\begin{figure*}[t] 
    \centering
    \begin{subfigure}{0.25\linewidth}
      \centering
      \includegraphics[width=1\linewidth, trim={0 0.5cm 0 0.5cm},clip]{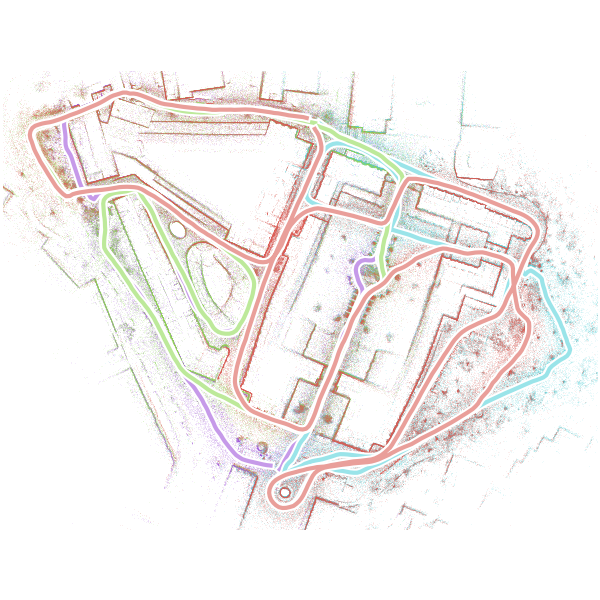}
      \vspace{-0.7cm}
      \caption{\texttt{kth\_r4\_00\_wifi}}
      \label{fig:rimesa-experiments:cosmobench:qualitative:kth}
    \end{subfigure}%
    \begin{subfigure}{0.25\linewidth}
      \centering
      \includegraphics[width=1\linewidth,trim={0 0.5cm 0 0.5cm},clip]{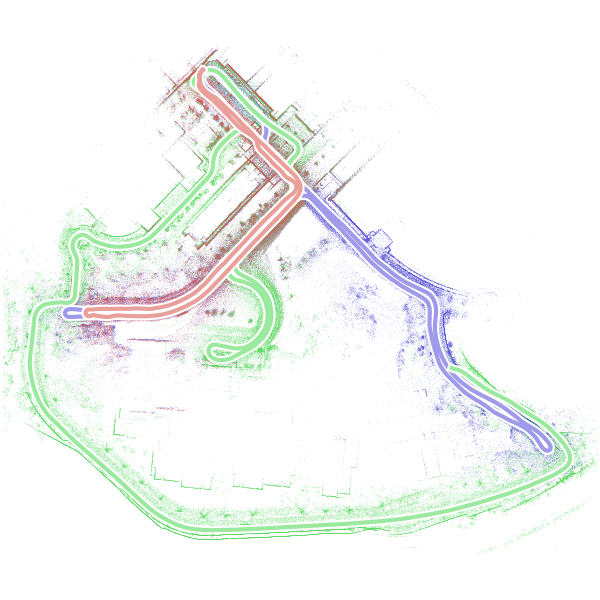}
      \vspace{-0.7cm}
      \caption{\texttt{ntu\_r3\_02\_proradio}}
      \label{fig:rimesa-experiments:cosmobench:qualitative:ntu}
    \end{subfigure}%
    \begin{subfigure}{0.25\linewidth}
      \centering
      \includegraphics[angle=270,width=1\linewidth,trim={0.5cm 0 0.5cm 0},clip]{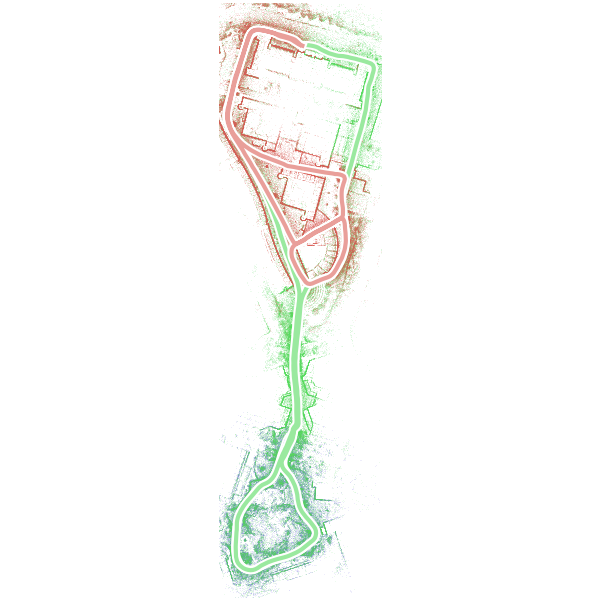}
      \vspace{-0.7cm}
      \caption{\texttt{tuhh\_r3\_01\_proradio}}
      \label{fig:rimesa-experiments:cosmobench:qualitative:tuhh}
    \end{subfigure}%
    \begin{subfigure}{0.25\linewidth}
      \centering
      \includegraphics[width=1\linewidth,trim={0 0.5cm 0 0.5cm},clip]{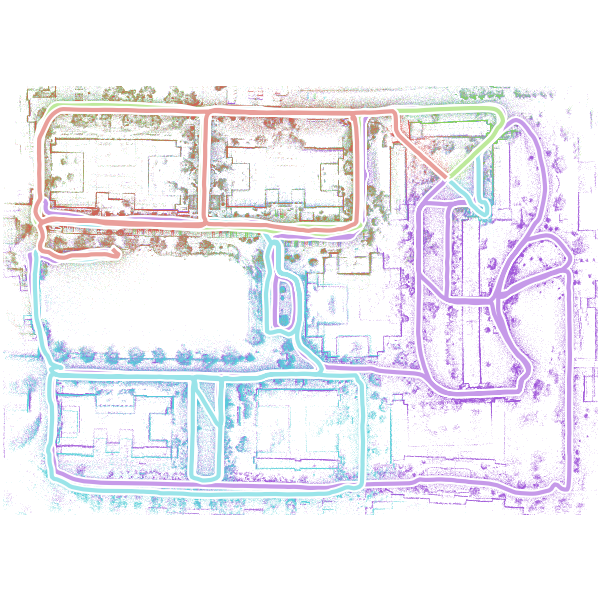}
      \vspace{-0.7cm}
      \caption{\texttt{main\_campus\_wifi}}
      \label{fig:rimesa-experiments:cosmobench:qualitative:cumulti}
    \end{subfigure}
    \vspace{-0.5cm}
    \caption{Qualitative results from riMESA on a sample of COSMO-Bench datasets. Colors represent different robots. The pointcloud map is constructed by transforming LiDAR scans according to the riMESA solution. Visible by the alignment between each robot's map, riMESA is able to provide accurate and consistent state estimates.}
    \label{fig:rimesa-experiments:cosmobench:qualitative}
    \vspace{-0.5cm}
\end{figure*}

\begin{table}[t]
    \centering
    \scriptsize
    \renewcommand*{\arraystretch}{1.2}
    \caption{iATE and Final ATE performance for all methods and baselines on the Nebula Datasets~\cite{lamp2_chang_2022} released by COSMO-Bench~\cite{cosmobench_mcgann_2025}. The best-performing method (excluding baselines) for each dataset is bolded. The proposed algorithm is able to closely match the performance of the Centralized GNC baseline (expected performance upper-bound), though all methods do generally well on these datasets.}
    \vspace{-0.2cm}
    \label{tab:rimesa-experiments:cosmobench:nebula:quantitative}
\newcolumntype{Y}{>{\centering\arraybackslash}X}
\begin{tabularx}{\linewidth}{|l|l|Y|Y|Y|Y||Y|Y|Y|Y|}
\hline
\multicolumn{2}{|c|}{Metric} & \multicolumn{4}{c||}{iATE (m)} & \multicolumn{4}{c|}{Final ATE (m)} \\
\hline
\multicolumn{2}{|c|}{Dataset} & \rotatebox[origin=c]{90}{\texttt{finals}} & \rotatebox[origin=c]{90}{\texttt{ku}} & \rotatebox[origin=c]{90}{\texttt{tunnel}} &  \rotatebox[origin=c]{90}{\texttt{urban}} & \rotatebox[origin=c]{90}{\texttt{finals}} & \rotatebox[origin=c]{90}{\texttt{ku}} & \rotatebox[origin=c]{90}{\texttt{tunnel}} &  \rotatebox[origin=c]{90}{\texttt{urban}} \\
\hline
\multirow{6}{*}[1.5ex]{\rotatebox[origin=c]{90}{Baselines}}
& \SymIndep~Independent &0.61 & 3.14 & 3.36 & 1.03 & 0.65 & 1.95 & 3.96 & 1.47\\
& \SymCentralized~Cent. Oracle &0.55 & 2.47 & 1.49 & 0.70 & 0.74 & 2.08 & 1.54 & 0.84\\
& \SymIMESA~iMESA &0.49 & 2.26 & 0.88 & 0.64 & 0.78 & 1.57 & 0.93 & 0.76\\
& \SymGNC~Cent. GNC &1.65 & 1.48 & 0.91 & 1.17 & 0.52 & 1.26 & 1.05 & 1.11\\
& \SymPCM~Cent. PCM &5.07 & 1.51 & 2.49 & 3.13 & 6.77 & 1.91 & 2.48 & 5.46\\
\hline
\multirow{4}{*}[0ex]{\rotatebox[origin=c]{90}{Methods}}
& \SymDDFSAM~DDF-SAM2 &6.74 & 1.43 & 2.78 & 7.03 & 17.1 & 1.30 & 2.88 & 12.0\\
& \SymDLGBP~DLGBP &\textbf{0.69} & 2.90 & \textbf{0.79} & 1.00 & 0.93 & 3.64 & 0.89 & 1.13\\
& \SymKIMESA~kiMESA &0.73 & 1.44 & 0.83 & \textbf{0.94} & \textbf{0.49} & 1.26 & 0.89 & \textbf{1.04}\\
&\cellcolor{gray!20} \SymRIMESA~riMESA &\cellcolor{gray!20} 1.71 &\cellcolor{gray!20} \textbf{1.37} &\cellcolor{gray!20} 0.85 &\cellcolor{gray!20} 1.06 &\cellcolor{gray!20} 0.66 &\cellcolor{gray!20} \textbf{1.20} &\cellcolor{gray!20} \textbf{0.87} &\cellcolor{gray!20} 1.11\\
\hline
\end{tabularx}

    \vspace{-0.1cm}
\end{table}

\subsubsection{Real-World Accuracy Results}\label{sec:rimesa-experiments:cosmobench:results}
\begin{figure}[t!]
    \centering
    \includegraphics[width=3.45in]{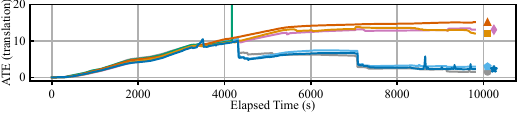}
    \caption{ATE plotted over time for \texttt{kth\_r3\_01\_wifi} for Centralized Oracle~(\SymCentralized), iMESA~(\SymIMESA), riMESA~(\SymRIMESA), kiMESA~(\SymKIMESA), DDF-SAM2~(\SymDDFSAM), DLGBP~(\SymDLGBP), and Independent~(\SymIndep). Unlike prior works that struggle to converge or diverge entirely (DDF-SAM2 at $s\approx 4200$), riMESA closely matches the Centralized Oracle across the entire trial.}
    \label{fig:rimesa-experiments:cosmobench:iate_vs_time}
    \vspace{-0.1cm}
\end{figure}

Comprehensive results for the COSMO-Bench datasets can be found in Tab.~\ref{tab:rimesa-experiments:cosmobench:main:quantitative}, and results from the Nebula datasets can be found in Tab.~\ref{tab:rimesa-experiments:cosmobench:nebula:quantitative}. For these results, we also provide the ATE for the final step of the dataset for both completeness and to provide better context to iATE values. Qualitative performance of riMESA can be seen in Fig.~\ref{fig:rimesa-experiments:cosmobench:qualitative} for a sample of the datasets, and an example plot of ATE over a single dataset can be seen in Fig.~\ref{fig:rimesa-experiments:cosmobench:iate_vs_time}.

\begin{remark}[Real-World Experiment Nuances]\label{remark:real-world-result-nuances}
While analyzing the real-world results keep in mind that our reference solutions, which are used to compute metrics, are not perfect. While derived using accurate additional sensors, these reference solutions do not necessarily represent true ``groundtruth.'' Secondly, due to real-world measurement noise, the global optimum of the C-SLAM problem does not necessarily correspond to either the groundtruth nor the reference solution. Combined, these realities mean that even our theoretical upper-bound methods (i.e. Centralized Oracle) may report metrics that appear worse than other methods. Despite this, these metrics still represent the practical usefulness of a solution; the trends, however, are noisier than those seen in synthetic data experiments.
\end{remark}

In these results we can see that riMESA is most frequently the best-performing method. There are some datasets for which riMESA is beaten by the ablative method kiMESA (e.g. \texttt{ntu\_r3\_01\_proradio}) or the prior works DDF-SAM2 or DLGBP (e.g. \texttt{ntu\_r4\_00\_wifi} and \texttt{finals}). However, unlike these methods, which also frequently produce poor results (e.g. \texttt{ntu\_r4\_00\_proradio}), riMESA is the only method that is able to perform consistently across all the datasets. Additionally, due to our use of incremental metrics, and as highlighted in Fig.~\ref{fig:rimesa-experiments:cosmobench:iate_vs_time}, we can see that riMESA performs consistently throughout robot operations.

To validate this high-level takeaway, we look at the average performance gap of the tested methods relative to the Centralized GNC baseline. Centralized GNC was selected as it is the best-performing baseline that could (in limited settings with reliable communication) be practically deployed (e.g. LAMP 2.0~\cite{lamp2_chang_2022}). The performance gap is computed between a method ($m$) and this baseline (GNC) as $[(\mathrm{iATE}_{m} - \mathrm{iATE}_{\mathrm{GNC}}) / \mathrm{iATE}_{\mathrm{GNC}}] \times 100$. This measure is averaged across all the real-world datasets, and for each method is:
\begin{minipage}{\linewidth} 
    \centering
    \footnotesize
    \setlength\tabcolsep{8pt}
    \renewcommand*{\arraystretch}{1.2}
    \vspace{4pt}
    \begin{tabular}{cccc}
        riMESA~\SymRIMESA & kiMESA~\SymKIMESA & DDF-SAM2~\SymDDFSAM & DLGBP~\SymDLGBP \\
        \hline
        45.09\% & 157.70\% & 787.30\% & 352.81\%
    \end{tabular}
    \vspace{4pt}
\end{minipage}
where we can see that riMESA's optimality gap is \textbf{>7x} lower than DLGBP and \textbf{>17x} lower than DDF-SAM2. Performance improvement over kiMESA is less significant, but recall this method is an ablative comparison for riMESA's robust local optimization. The performance gap relative to the baseline is non-trivial at 45\%; however, unlike the Centralized GNC baseline, riMESA must contend with limited communication between robots, and a performance drop is expected.

From this experiment we can derive some additional insights. Interestingly, we can see that Centralized PCM can struggle to produce quality results on some datasets. This poor performance is likely caused by PCM misclassifying outlier measurements, which are then incorporated into a non-robust optimizer producing unusable solutions. This provides additional evidence for the use of continuation-based approaches like riMESA. We can also see that the most challenging datasets appear to be \texttt{kittredge\_loop} and \texttt{main\_campus} (Tab.~\ref{tab:rimesa-experiments:cosmobench:main:quantitative}). Notably, for these datasets there is a larger gap between the performance of the ``oracle'' baselines and any robust method. This indicates that these datasets are challenging due to the presence of outliers, though their long duration is likely a compounding factor. On the other hand, the ``easiest'' datasets appear to be the Nebula datasets (Tab.~\ref{tab:rimesa-experiments:cosmobench:nebula:quantitative}) on which most methods perform reasonably. This is likely due to these datasets containing accurate odometry from a system that was fine-tuned for the operation environment~\cite{lamp2_chang_2022}.

\begin{figure}[t] 
    \begin{subfigure}{1.0\linewidth}
        \centering
        \includegraphics[width=1.75in]{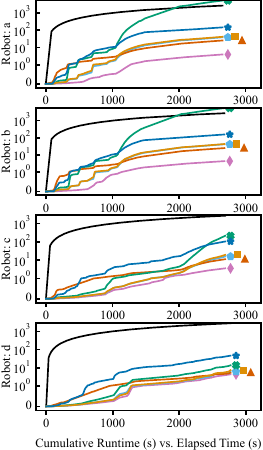}%
        \includegraphics[width=1.68in]{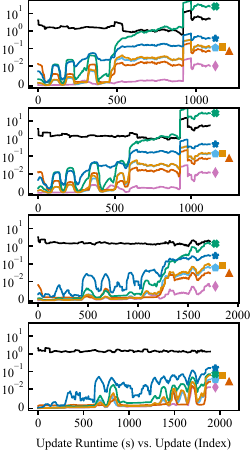}  
        \vspace{-0.1cm}
        \caption{Distributed Algorithm Timing}
        \vspace{4pt}
        \label{fig:rimesa-experiments:cosmobench:timing:distributed}
    \end{subfigure}
    \begin{subfigure}{1.0\linewidth}
        \centering
        \includegraphics[width=3.45in]{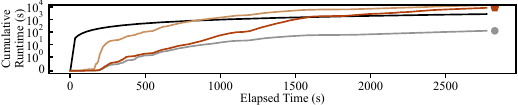}%
        \vspace{4pt}
        \includegraphics[width=3.45in]{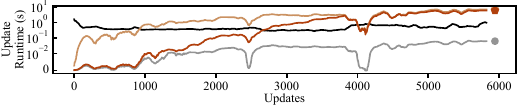}
        \vspace{-0.4cm}
        \caption{Centralized Algorithm Timing}
        \label{fig:rimesa-experiments:timing:centralized}
    \end{subfigure}
    \vspace{-0.45cm}
    \caption{The runtime performance (cumulative and per-update) on \texttt{main\_campus\_wifi}. (a) Distributed algorithm performance -- iMESA~(\SymIMESA), riMESA~(\SymRIMESA), kiMESA~(\SymKIMESA), DDF-SAM2~(\SymDDFSAM), DLGBP~(\SymDLGBP), and Independent~(\SymIndep). (b) Centralized baseline performance --  Centralized Oracle~(\SymCentralized), Centralized GNC~(\SymGNC), and Centralized PCM~(\SymPCM). The real-time bounds for the dataset are shown by black lines ($\strns$) and is not constant for updates as time between measurements varies. Apart from DDF-SAM2, all distributed methods are able to maintain real-time performance. While the Centralized Oracle can maintain real-time performance, the robust centralized methods required for real-world operation are too computationally expensive to operate in real-time. \textit{Note: The y-axis scales are linear up to 0.01s and 1s for update/cumulative runtime plots, respectively, and log-scale above these thresholds.}}
    \label{fig:rimesa-experiments:cosmobench:timing}
    \vspace{-0.1cm}
\end{figure}

\subsubsection{Real-World Timing Results}\label{sec:rimesa-experiments:cosmobench:timing}
A key challenge of our target problem is that robots operate online and require up-to-date state estimates for downstream tasks. We next evaluate the runtime of riMESA using the real-world COSMO-Bench datasets. We focus on \texttt{main\_campus\_wifi} as its long duration provides a quality stress test for runtime performance. 

We use a machine with an Intel Core i9-13900K processor and 128 GB of RAM. Each method was run independently, and care was taken to minimize background processes. We look at both update runtime and total cumulative runtime. To support practical deployments, a method must remain below real-time bounds for both of these measures. We plot the runtimes for all methods in Fig.~\ref{fig:rimesa-experiments:cosmobench:timing}. For distributed algorithms that support parallel computation, we plot the runtime per-robot, for the centralized algorithms, we plot results for the central server.

In Fig.~\ref{fig:rimesa-experiments:cosmobench:timing} we can see that almost all the distributed algorithms are able to achieve real-time performance. The exception is DDF-SAM2, which struggles with long update runtimes towards the end of the dataset. These slow updates are likely caused by the algorithm's computation of marginals. On the other hand, while the Centralized Oracle is able to maintain real-time performance, both robust centralized algorithms exceed real-time bounds. While there are efficiency improvements that could be implemented to reduce the runtimes of these algorithms (Remark~\ref{remark:centralized-baseline-operation}), the aggregation of all data simply results in an intractably large problem for these robust centralized methods, even for the relatively small team size of 4 robots that was tested in this experiment.

\begin{remark}[Centralized Baseline Operation]\label{remark:centralized-baseline-operation}
    Centralized algorithms are updated for every new measurement. Odometry is integrated into existing solutions, and loop-closures cause batch re-optimization using the previous solution as a warm start. Due to asynchronous measurements from each robot, centralized algorithms perform more updates than individual robots in the distributed case. Practical implementations could aggregate measurements and perform fewer optimization processes. While this would reduce the cumulative runtime of the method, it would increase the per-update runtime and delay new information from being utilized by the team. 
\end{remark}

\subsubsection{Real-World Bandwidth} \label{sec:rimesa-experiments:cosmobench:bandwidth}
As noted, across these experiments we simulate unreliable connectivity, limited communication range, and two-general's failures between robots. However, we did not limit communications based on bandwidth. In this section we evaluate the validity of this assumption. For all trials, we record the total bandwidth usage for the entire team each second. Across all trials using the Wi-Fi model, riMESA had a peak bandwidth usage of $625.3$ KB/s and an average usage of $54.9$ KB/s. Across all trials using the Pro-Radio model, including the Nebula dataset trials, riMESA had a peak usage of $825.3$ KB/s and an average usage of $57.6$ KB/s.

While these results are conditioned on the communication model and robot trajectories, they do validate our experimental assumption, as riMESA's bandwidth usage remains below the limits of the Wi-Fi ($2000$ KB/s) and Pro-Radio ($1000$ KB/s) models. It is also worth noting that these are conservative numbers, as communication pairs sufficiently far from each other will not interfere, making the practical bandwidth higher. Finally, for applications where bandwidth is extremely limited, communications could be throttled to operate within available bounds. Moreover, our results support riMESA's efficacy under such conditions, like the \texttt{ntu\_r3\_02\_wifi} trial in which riMESA had a peak bandwidth usage of only $3.8$ KB/s and an average usage of only $0.12$ KB/s and still outperformed all prior works and even some centralized methods.
\section{Conclusion}\label{sec:conclusion}
In this paper, we presented Consensus Alternating Direction Method of Multipliers as a framework to design effective C-SLAM back-ends. We then proposed riMESA, a robust, incremental, and fully distributed C-SLAM back-end that can handle the challenging conditions of real-world deployments and collaboratively produce high quality state estimates for multi-robot teams. riMESA incrementalizes an edge-based C-ADMM process to allow robots to operate incrementally and asynchronously, leverages a robust incremental local solver (riSAM) to address outlier measurements and compute updates in real-time, and utilizes a robust communication procedure to tolerate unreliable communication with latency. We evaluate the performance of riMESA under a variety of conditions on a total of 2430 unique synthetic datasets and on a total of 28 real-world datasets -- validating that riMESA consistently produces high-quality state estimates for a multi-robot team in real-time with only limited communication.

\section{Acknowledgments}\footnotesize
This work was partially supported by NASA award 80NSSC24CA020 and the NSF Graduate Research Fellowship Program.

\vspace{-0.2cm}
{
\bibliographystyle{IEEEtran}
\scriptsize
\bibliography{refs}
}

\vspace{-0.1cm}
\footnotesize%
\begin{IEEEbiography}[{\includegraphics[width=1in,height=1.25in,clip,keepaspectratio]{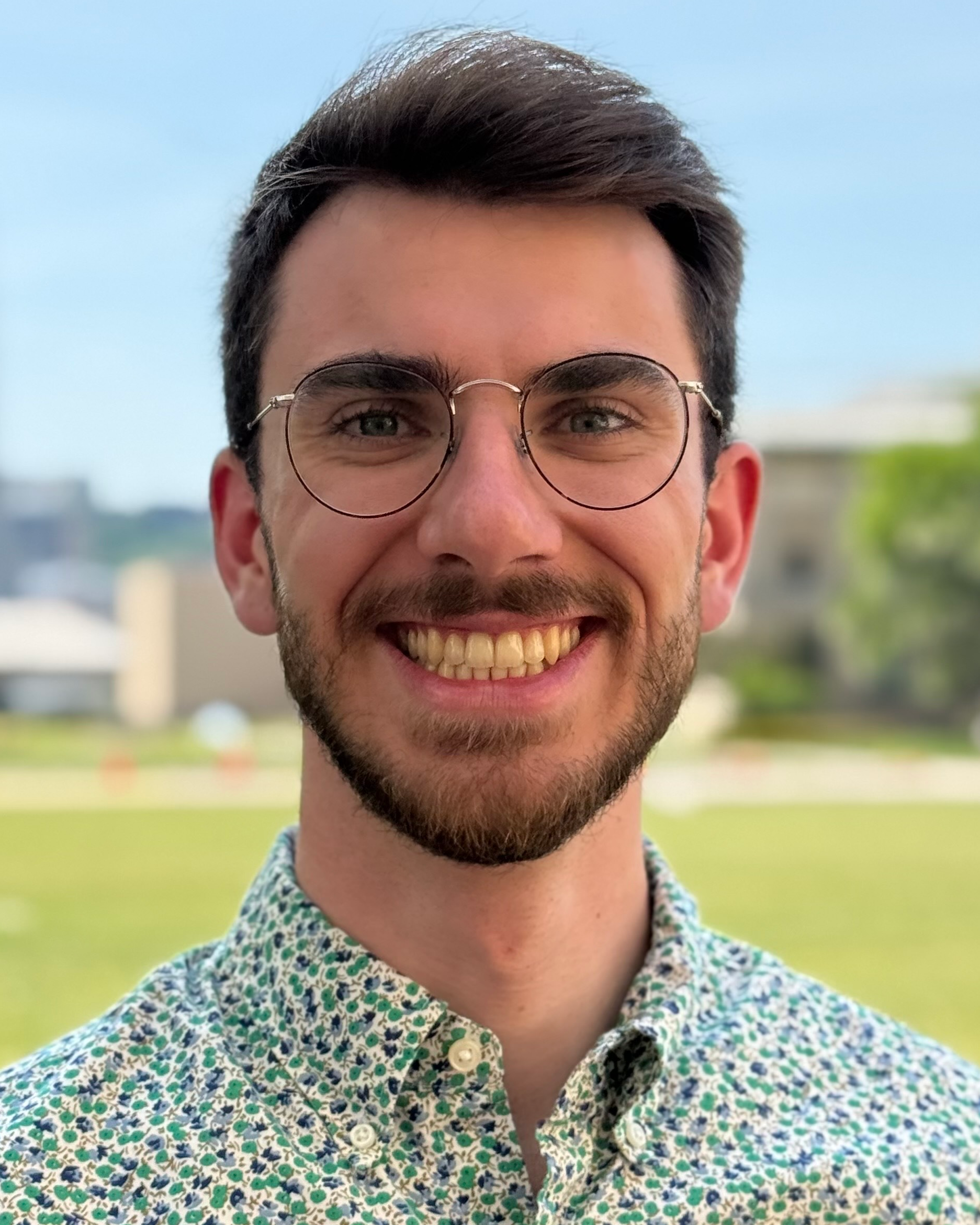}}]{Daniel McGann} received his B.S. in Computer Science from Northeastern University, Boston, MA, USA, in 2020 and his Ph.D. in Robotics from Carnegie Mellon University, Pittsburgh, PA, USA, in 2025. His research to date has focused on distributed optimization to enable Simultaneous Localization and Mapping (SLAM) for multi-robot teams. He is broadly interested in research to develop reliable state estimation and perception for robots and multi-robot teams operating in the wild.
\end{IEEEbiography}

\vspace{-0.34cm}
\footnotesize%
\begin{IEEEbiography}[{\includegraphics[width=1in,height=1.25in,clip,keepaspectratio]{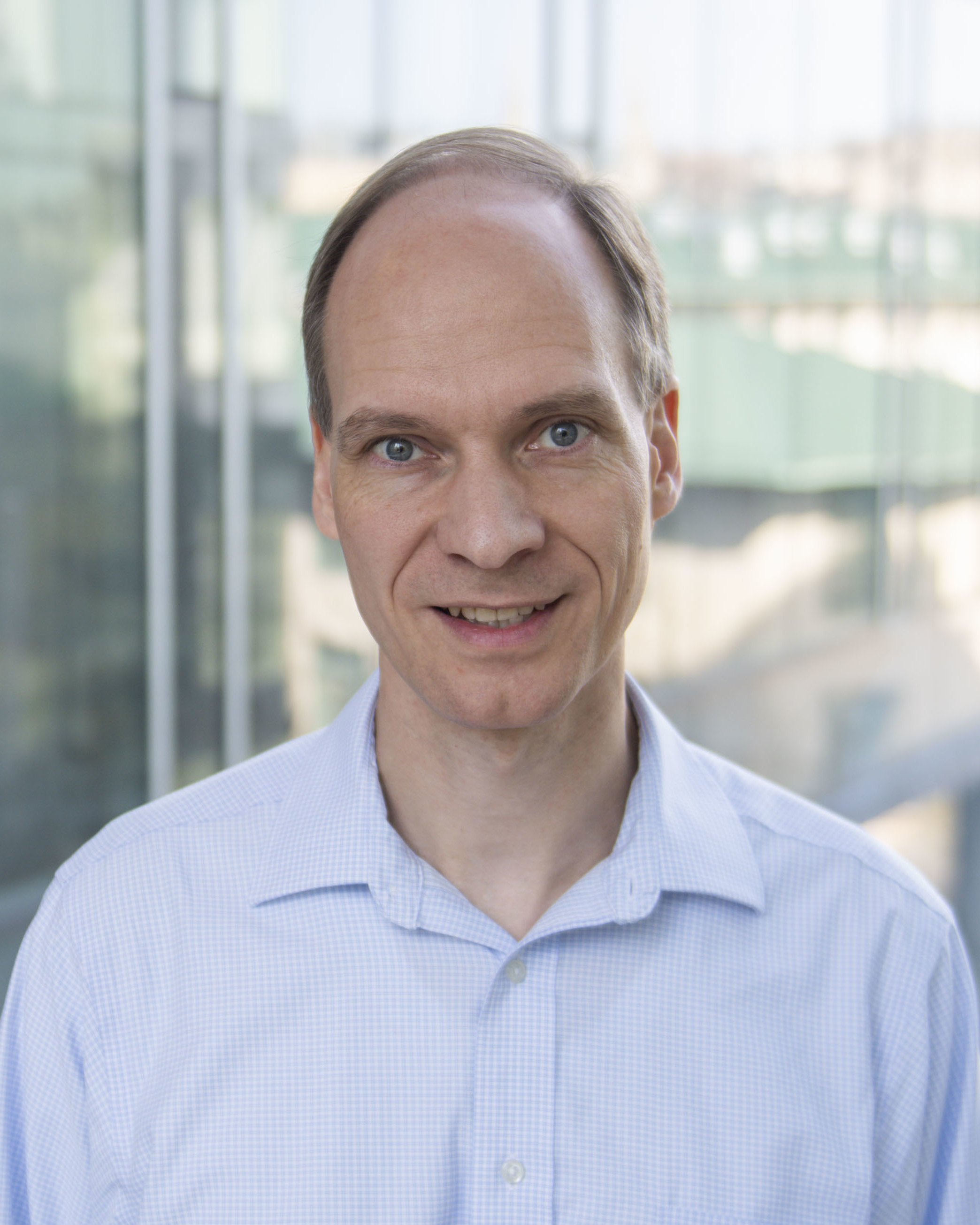}}]{Michael Kaess} is a Professor in the Robotics Institute at the School of Computer Science, Carnegie Mellon University (CMU). Before joining CMU in 2013, he was a Research Scientist and Postdoctoral Associate at the Computer Science and Artificial Intelligence Laboratory (CSAIL) at the Massachusetts Institute of Technology (MIT). He earned his PhD in Computer Science from the Georgia Institute of Technology in 2008. His research focuses on probabilistic methods for robot perception, including navigation, mapping, localization, and inference. 
\end{IEEEbiography}

\end{document}